\documentclass[10pt,journal,compsoc]{IEEEtran}

\ifCLASSOPTIONcompsoc
  \usepackage[nocompress]{cite}
\else
  \usepackage{cite}
\fi

\ifCLASSINFOpdf
\else
\fi

\usepackage{times}
\usepackage{epsfig}
\usepackage{graphicx}
\usepackage{amsmath}
\usepackage{amssymb}
\usepackage{multirow}

\usepackage[numbers,sort]{natbib} 
\usepackage{tabularx} 
\usepackage{booktabs} 
\usepackage{comment} 
\usepackage{color}
\usepackage{xcolor}
\usepackage{pgf,pgffor,pdfpages}
\usepackage{comment} 

\usepackage{enumerate}
\usepackage{enumitem}

\usepackage{xspace}

\usepackage{}

\usepackage{float}
\usepackage{soul}

\usepackage{pifont}
\usepackage{caption}

\usepackage[pagebackref=true,breaklinks=true,letterpaper=true,colorlinks,bookmarks=false]{hyperref}

\newcommand{\eg}{e.g.\@\xspace} 

\newcommand{\ie}{i.e.\@\xspace}

\newcommand{\etc}{etc.\@\xspace}

\newcommand{\hidcit}[1]{\ifx #1 \fi}
\newcommand{\hidcitb}[1]{\ifx #1 \fi}

\definecolor{amber}{rgb}{1.0, 0.49, 0.0}

\newcommand{\major}[1]{\textcolor{black}{#1}}

\newcommand{\UCOchance}{40.4}
\newcommand{\UCOLAEO}{UCO-LAEO\xspace}
\newcommand{\AVAchance}{17.1}
\newcommand{\AVALAEOscore}{39.1}
\newcommand{\AVALAEOscoreTrAVA}{50.6}
\newcommand{\UCOLAEOscore}{79.5}
\newcommand{\UCOLAEOscoreTrAVA}{77.8}
\newcommand{\TVHIDscoreTrAVA}{90.7}

\newcommand{\UCOLAEOscoreTrUCOSSMA}{81.5} 
\newcommand{\UCOLAEOscoreTrUCOSSMB}{81.3} 
\newcommand{\UCOLAEOscoreTrUCOSSMC}{86.7} 
\newcommand{\AVALAEOscoreTrAVASSMA}{59.8} 
\newcommand{\AVALAEOscoreTrAVASSMB}{68.4} 
\newcommand{\AVALAEOscoreTrAVASSMC}{68.7} 
\newcommand{\TVHIDscoreBestTrUCO}{92.3}
\newcommand{\TVHIDscoreBestTrAVA}{87.4}
\newcommand{\AVALAEOscoreTrUCOFtAVA}{67.0}
\newcommand{\UCOLAEOscoreTrAVAFtUCO}{84.5}
\let\oldsim\sim 
\renewcommand{\sim}{{\oldsim}}
\newcommand{\newnet}{LAEO-Net++\@\xspace}
\newcommand{\oldnet}{LAEO-Net\@\xspace}

\renewcommand{\paragraph}[1]{\vspace{4mm} \noindent \textbf{#1}}

\newcommand\beforecaptions{\vspace{-2mm}}
\newcommand\aftercaptions{\vspace{-2mm}}

\begin{document}

\title{\newnet: revisiting people Looking At Each Other in videos}

\author{Manuel J. Mar\'in-Jim\'enez$^*$, Vicky Kalogeiton$^*$, Pablo Medina-Su\'arez, and Andrew Zisserman%
\IEEEcompsocitemizethanks{
\vspace{-3mm}
\IEEEcompsocthanksitem ($^*$) means equal contribution.
\IEEEcompsocthanksitem Manuel J. Mar\'in-Jim\'enez and Pablo Medina-Su\'arez are with the University of C\'ordoba, Spain. 
Emails: \href{mailto:mjmarin@uco.es}{\textcolor{black}{mjmarin@uco.es}} and \href{mailto:i42mesup@uco.es}{\textcolor{black}{i42mesup@uco.es}}
\IEEEcompsocthanksitem Vicky Kalogeiton is at LIX, École Polytechnique and Andrew Zisserman at the University of Oxford. \protect
Emails: \href{mailto:vicky.kalogeiton@polytechnique.edu}{\textcolor{black}{vicky.kalogeiton@polytechnique.edu}} and \href{mailto:az@robots.ox.ac.uk}{\textcolor{black}{az@robots.ox.ac.uk}}}%
}

\markboth{Journal of \LaTeX\ Class Files,~Vol.~14, No.~8, August~2020}%
{Shell \MakeLowercase{\textit{et al.}}: Bare Demo of IEEEtran.cls for Computer Society Journals}


\IEEEtitleabstractindextext{%
\begin{abstract}
Capturing the `mutual gaze' of people is essential for understanding and interpreting the social interactions between them. 
To this end, this paper addresses the problem of detecting people \textit{Looking At Each Other (LAEO)} in video sequences. 
For this purpose, we propose \newnet, a new deep CNN for determining LAEO in videos. 
In contrast to previous works, \newnet takes spatio-temporal tracks as input and reasons about the whole track. 
It consists of three branches, one for each character's tracked head and one for their relative position. 
Moreover, we introduce two new LAEO datasets: \UCOLAEO and AVA-LAEO.
A thorough experimental evaluation demonstrates the ability of \newnet to successfully determine if two people are LAEO and the temporal window where it happens. 
Our model achieves state-of-the-art results on the existing TVHID-LAEO video dataset, significantly outperforming previous approaches. 
Finally, we apply \newnet to a social network, where we automatically infer the social relationship between pairs of people based on the frequency and duration that they LAEO, and show that LAEO can be a useful tool for guided search of human interactions in videos. The code is available at \url{https://github.com/AVAuco/laeonetplus}.
\end{abstract}

}

\maketitle
\IEEEdisplaynontitleabstractindextext

%
\IEEEpeerreviewmaketitle

\IEEEraisesectionheading{
\section{Introduction}
\label{sec:intro}
}

\IEEEPARstart{E}{ye} contact or `mutual gaze' is an important part of the non-verbal
communication between two people~\cite{loeb1972mutual}.  The duration and frequency
of eye contact depends on the nature of the relationship and reflects
the power relationships, the attraction or the antagonism between the
participants~\cite{abele1986gaze}.  
Therefore, in order to understand and interpret the social interactions that are occurring, it is important to capture this signal accurately. 
The importance of detecting people Looking At Each Other (LAEO) has already been recognized in a series of computer vision papers~\cite{marin2013ijcv,palmero2018laeo}  as well as in other papers that study human gaze~\cite{chong2018eccv,recasens2015nips,recasens2017iccv,brau2018eccv}.

\begin{figure*}[t]
\centerline{
\includegraphics[width=1\linewidth]{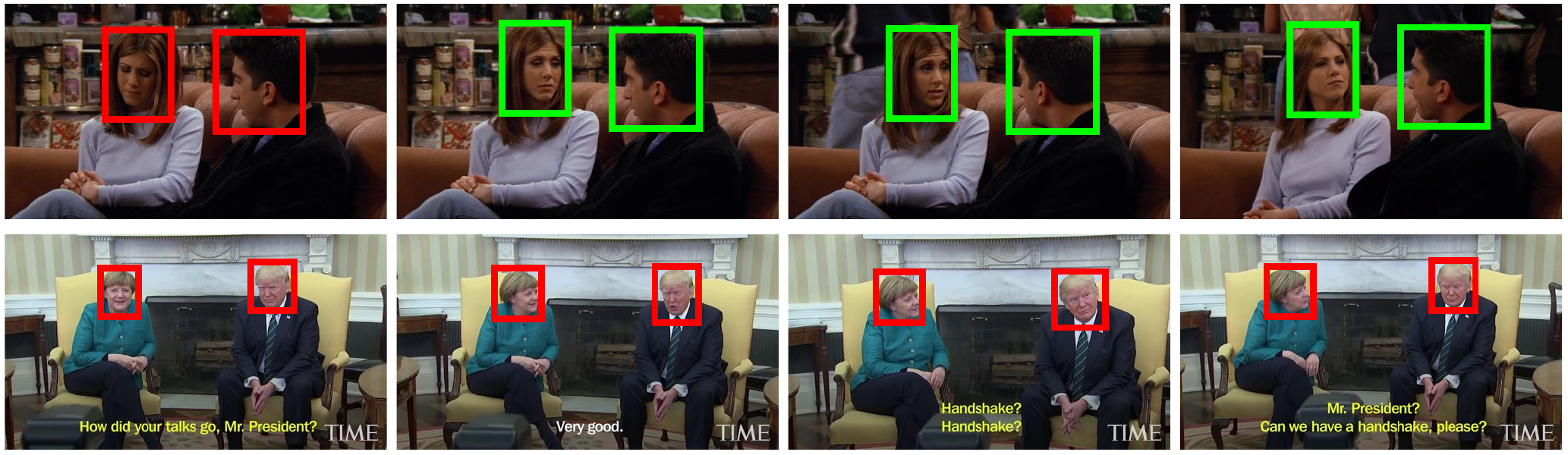}
}
 \caption{\small{\textbf{Intimacy or hostility?} Head pose, along with body pose and facial expressions, is a rich source of information for interpreting human interactions. Being able to automatically understand the non-verbal cues provided by the relative head orientations of people in a scene enables a new level of human-centric video understanding. Green and red pairs of heads represent LAEO and non-LAEO cases, respectively. Video source of second row: \url{https://youtu.be/B3eFZMvNS1U}
   }}
\aftercaptions
 \label{fig:teaser}
\end{figure*}

LAEO is complementary to other forms of human non-verbal communication such as facial expressions, 
gestures, proxemics (distance), body language and pose, paralanguage (the tone of the voice, prosody), and 
interactions (e.g.\ hugging, handshake). Many of these have been the subject of recent papers \cite{marin2014mva,vondrick2016cvpr,gu2018ava,kukleva2020learning}. 
In this paper, we introduce a new deep convolutional neural network (CNN) for determining LAEO in video
material, coined \textbf{\newnet}. Unlike previous works, our approach answers the question of whether two characters are LAEO over a temporal period by using a spatio-temporal model, whereas previous models have only considered individual frames.  
The problem with frame-wise LAEO is that when characters blink or momentarily move their head, then they are considered non-LAEO, and this can severely affect the accuracy of the LAEO measurement over a time period.
The model we introduce considers head tracks over multiple frames, and determines whether two characters are LAEO for a time period based on the pose of their heads and their relative position. Such an example is in Figure~\ref{fig:teaser}.

We make the following contributions: first, we introduce a spatio-temporal LAEO model that consists of three branches, one for each character's tracked head and one for their relative position, 
together with a fusion block. This is described in Section~\ref{sec:model}. 
To the best of our knowledge, this is the first work that uses tracks as input and reasons about people LAEO in the whole track, instead of using only individual frames.
Second, we introduce two new datasets (Section~\ref{sec:datasets}): 
(i)~{\bf \UCOLAEO}, a new dataset for training and testing LAEO. It consists of  $129$ ($3$-$12$ sec) clips from four popular TV shows; and 
(ii)~{\bf AVA-LAEO}, a new dataset, 
which extends the existing large scale AVA dataset~\cite{gu2018ava} with LAEO annotations for the training and validation sets.  
We evaluate the performance of the spatio-temporal LAEO model on both these new datasets (Section~\ref{sec:expers}).
Third, we show that our model achieves the state of the art on the existing TVHID-LAEO dataset~\cite{marin2013ijcv} by a significant margin ($3\%$). 
Finally, in Section~\ref{sec:friends}, we 
show that the LAEO score can be used as tool not only for demonstrating social relationships between people but also for guiding the search for human interactions in videos; we demonstrate these for one episode of the TV comedy `Friends'.

A preliminary version of this work has been published in CVPR 2019~\cite{marin19cvpr}. We significantly extend it in the following ways:
\begin{itemize}
    \item \textbf{Design.} We propose \newnet: a new three branch head-track model for determining if two people are LAEO. 
    \newnet is based on \oldnet~\cite{marin19cvpr} but it better decodes the head-tracks by using a different architecture and it better exploits the temporal continuity of videos by using $\mathcal{M}$-frames long head-tracks. 
    The differences between the two models are described in details in Section~\ref{sec:model} and experimentally compared in Section~\ref{sub:old_vs_new}. 
    The results show that the proposed changes improve the performance and, overall, \newnet outperforms all other approaches. 
    \item \textbf{Pre-training schemes.} 
    We present three different settings with different levels of supervision to pre-train \newnet and discuss our findings (Section~\ref{sub:head-pose}). 
    First, we use ground truth labels for head orientation in videos (fully supervised setting). 
    Second, we use the self-supervised Facial Attributes-Net~\cite{wiles2018bmvc} by extending it to video frames 
    (self-supervised setting). 
    Third, we use random initialization and demonstrate that pose can be learnt implicitly by the LAEO task alone (implicit supervised setting).  
    \item \textbf{Analysis and experiments.} We provide more insights and contents to explain the performance of \newnet, as well as more experiments on all datasets (Sections~\ref{sub:head-map}-\ref{sub:cross}). 
    Specifically, we experimentally demonstrate and discuss the benefits of the head-map branch, exploiting the temporal dimension, implicit or explicit self-supervised learning, and of applying \newnet to various datasets (Section~\ref{sub:discussion}).    
    \item \textbf{Interaction prediction.} For one episode of the TV show `Friends' we use \newnet as a proxy for guiding the search for human interactions in videos. In particular, we show that by using LAEO we can identify the social relationship between characters and whether two characters are interacting, even if they hardly co-exist (Section~\ref{sec:friends}).
\end{itemize}

\section{Related work} 
\label{sub:relworks}

Gaze~\cite{recasens2015nips} and head pose~\cite{drouard2017hpose} are powerful tools to deal with the problem of determining the \textit{visual focus of attention} (VFoA) in a scene, \ie what people are looking at. 
For instance, \cite{kobayashi2001unique} highlights the importance of the white part of the human eye (\ie white sclera) in recognising gaze direction, enabling the extraordinary ability of humans to communicate just by using gaze signals.

\paragraph{Visual focus of attention.} 
One classical approach for determining the VFoA is \cite{ba2009vfoa}, where the authors model the dynamics of a meeting group in a probabilistic way, inferring where the participants are looking at. 
An improved version of this work is presented in \cite{ba2011pami}, where context information is used to aid in solving the task. 
\cite{zhang17pami} present a new gaze dataset and propose GazeNet, the first deep appearance-based gaze estimation method. 
More recently, \cite{brau2018eccv} discover 3D locations of regions of interest in a video by analysing human gaze. 
They propose a probabilistic model that simultaneously infers people's location and gaze as well as the item they are looking at, which might even be outside the image.

\begin{figure*}[t]
\centerline{
\includegraphics[width=1\linewidth]{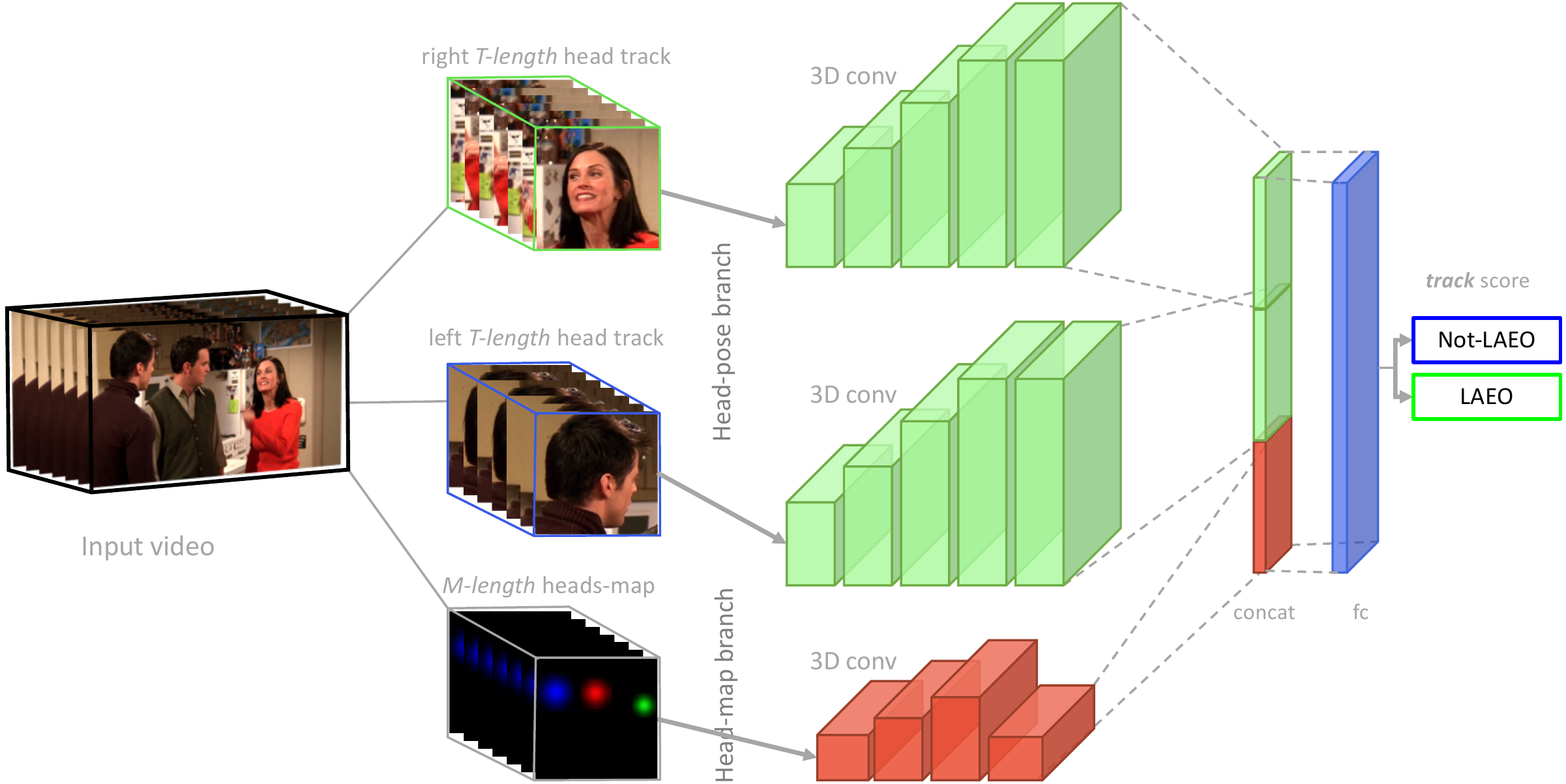}
}
 \caption{\small{\textbf{Our three branch track \newnet}: It consists of the head branches (green), the head-map branch (red) and a fusion block, which concatenates the embeddings from the other branches and scores the track sequence as LAEO or not-LAEO with a fully connected layer (blue) using softmax loss. In our experiments, we use head tracks of length $\mathcal{T}=10$ and head-maps of length $\mathcal{M}=10$}.}
\label{fig:main}
\end{figure*}

\paragraph{Gaze direction.}
In the literature, some works focus on `gaze following'~\cite{recasens2015nips,recasens2017iccv}. 
\cite{recasens2015nips} proposes a two-branch model that follows the gaze of a single person (head branch) and identifies the object being looked at (saliency branch). 
In a similar manner, 
\newnet makes use of spatial and temporal information throughout the video and processes the relation of people over time. We discuss the relationship between LAEO and gaze prediction learning in Section~\ref{sub:pre-training}.

The work in \cite{chong2018eccv} focuses on images by proposing a network that estimates both the gaze direction and the VFoA. A coarse spatial location of the target face is provided in the form of a one-hot vector. In contrast, in our model, this is provided by a RGB image with Gaussian-like circles representing the centre and scale of heads and a colour-coding indicating the target pair (Figure~\ref{fig:headmaps_synpairs}~(a)). Thus, our representation offers a better resolution of the scene geometry and incorporates cues about head scales.

Typically, in edited movies, an interaction is represented by alternating video shots. Therefore, sometimes the VFoA is not visible in the current frame or shot, but in a different one. This is addressed in~\cite{recasens2017iccv} with a model that reasons about human gaze and 3D geometric relations between different views of the same scene. 
\cite{masse2018pami} consider scenarios where multiple people are involved in a social interaction. 
Given that the eyes of a person are not always visible (\eg due to camera viewpoint), they estimate people's gaze by modelling the motion of their heads with a Bayesian model.

\cite{fischer18eccv} propose an appearance-based CNN that learns a direct image-to-gaze mapping using a large dataset of annotated eye images. 
\cite{krafka2016cvpr} present GazeCapture, a dataset collected from smartphone users, and use it to train iTracker, a CNN for gaze estimation that runs in real-time on commercial mobile devices.
\cite{huang17mva} work on gaze estimation on tablets. 
They collect an unconstrained dataset
and present an method for gaze estimation using a Random Forest regressor. 
\cite{zhu17iccv} propose a gaze transform layer to connect separate head pose and eyeball movement models. This does not suffer from overfitting of head-gaze correlation and makes it possible to use datasets existing for other tasks.

\cite{li18eccv} propose a model for joint gaze estimation and action recognition in first person. 
They model gaze distribution using stochastic units, from which they generate an attention map. Then, this map guides the aggregation of visual features for action recognition. 
\cite{kellnhofer19iccv} collect a 3D gaze dataset simply by recording with an omni-directional camera subjects looking at a pre-defined point (indoors and outdoors).

\paragraph{People Looking At Each Other. }
A special case of VFoA is when subject-A's VFoA is subject-B, and subject-B's VFoA is subject-A. This is known as \textit{mutual gaze} or people \textit{looking at each other} (LAEO). 
This situation typically entails non-physical human interactions, but might precede or continue a physical one, \eg hand-shake before or after a conversation. 
In the context of \textit{Behaviour Imaging}, detecting LAEO events is a key for understanding higher-level social interactions, as in autism in children~\cite{rehg2011behavior}. 
Furthermore, \cite{ajodan2019increased} shows that children diagnosed with autism spectrum disorder demonstrate increased eye contact with their parents compared to others, \eg a clinician, despite the social communication difficulties. 
In the context of {social interaction}, \cite{goffman2008public,loeb1972mutual} point out that one principal way of demonstrating interest in social interaction is the willingness of people to LAEO.

The problem of detecting people LAEO in videos was introduced in \cite{marin2013ijcv}. After detecting and tracking human heads,  \cite{marin2013ijcv} model and predict yaw and pitch angles with a Gaussian Process regression model. 
Based on the estimated angles and the relative position of the two heads, a LAEO score is computed per frame, and aggregated over the shot. 
Although we also model the head pose and relative position, \newnet estimates LAEO for a track over a temporal window, instead of a single frame.

\cite{ricci2015iccv} address the problem of detecting conversational groups in social scenes by combining cues from body orientation, head pose and relative position of people. 
In a controlled scenario with just two people, \cite{palmero2018laeo} addresses the LAEO problem by using two calibrated cameras placed in front of the participants, making sure that there is an overlapping visible zone between both cameras. 
Recently, LAEO has been used as an additional task in the joint learning of LAEO and 3D gaze estimation~\cite{doosti2020mgaze3d}. 
The authors show that this leads to richer representations than solving each task separately, and more importantly that 3D gaze is a powerful cue for understanding relations.  
Thus LAEO is bridging the gap between 2D and 3D mutual gaze detection (more details in Section~\ref{sub:pre-training}).

\paragraph{Interactions and relations. }
Looking at a person is a dominant classes for human interactions in videos~\cite{Patron2010hi5,gu2018ava}. 
~\cite{liu2019social} propose a network to capture long and short-term temporal cues, 
\cite{lv2018multi} classify relationships between characters, while \cite{kukleva2020learning} jointly learn interactions and relations between characters. 
Instead, we treat mutual gaze as a cue to identify the existence of interactions and to determine the level of friendness between people. 
In this context, we think that a LAEO model (either pre-trained or fine-tuned and adapted to new data) can have an impact on other applications, such as detecting cartoons, animals (\eg cats, chimpanzees~\cite{schofield19chimp}) or other object classes (\eg cars) looking at each other.

\section{\newnet }
\label{sec:model}

\begin{table*}[ht!]
\begin{minipage}[b]{0.7\textwidth}
\centering
    \small{
    \begin{tabular}{ll rr rr r}
    \toprule
    
    \multicolumn{2}{l}{\multirow{2}{*}{statistics}}& \multicolumn{2}{c}{\UCOLAEO (\textbf{new})} & \multicolumn{2}{c}{AVA-LAEO (\textbf{new})} &
    \multicolumn{1}{c}{TVHID-LAEO~\cite{marin2013ijcv}} \\
    
    & & \multicolumn{1}{c}{train+val} & \multicolumn{1}{c}{test} & \multicolumn{1}{c}{train} & \multicolumn{1}{c}{val} & \multicolumn{1}{c}{test} \\
    \midrule
    \multicolumn{2}{l}{\#frames} & \multicolumn{2}{c}{$>18$k} & \multicolumn{2}{c}{$>1,4$M (estim.)} & \multicolumn{1}{c}{$>29$k} \\ 
    \multicolumn{2}{l}{\#programs} & \multicolumn{2}{c}{4 (tv-shows)} & \multicolumn{2}{c}{298 (movies)} & \multicolumn{1}{c}{20 (tv-shows)}\\
    \multirow{2}{*}{shots} & ~~~~\#annotations & 106+8 & 15  &  40166 & 10631  & 443  \\ 
    & ~~~~\#LAEO & 77+8 & 15 & 18928 & 5678  & 331  \\ 
    \multirow{2}{*}{pairs} & ~~~~\#annotations & 27358+5142 & 3858 & 137976 & 34354  & -- \\ 
    & ~~~~\#LAEO & 7554+1226 & 1558 & 19318 & 5882  & -- \\ 
    
    \multicolumn{2}{l}{sets (pairs)} & 32500 & 3858 & 137976 & 34354 & 443  (shots)\\
    \bottomrule
    \end{tabular}
    }
    \captionof{table}{\small{\textbf{Summary of LAEO datasets.}   \textit{\#programs}:  different TV shows; \textit{\#shot-annotations}: annotated shots; \textit{\#shot-LAEOs}: shots containing at least one LAEO pair; \textit{\#pair-annotations}: annotated human bounding box pairs; \textit{\#pair-LAEOs}: human bounding box pairs that are LAEO; \textit{sets:} \#train/val/test LAEO pairs (or shots) used.}}
    \label{tab:dbstats}
\end{minipage}
\hfill
\begin{minipage}[b]{0.29\textwidth}
    \centering
    \includegraphics[width=0.87\textwidth]{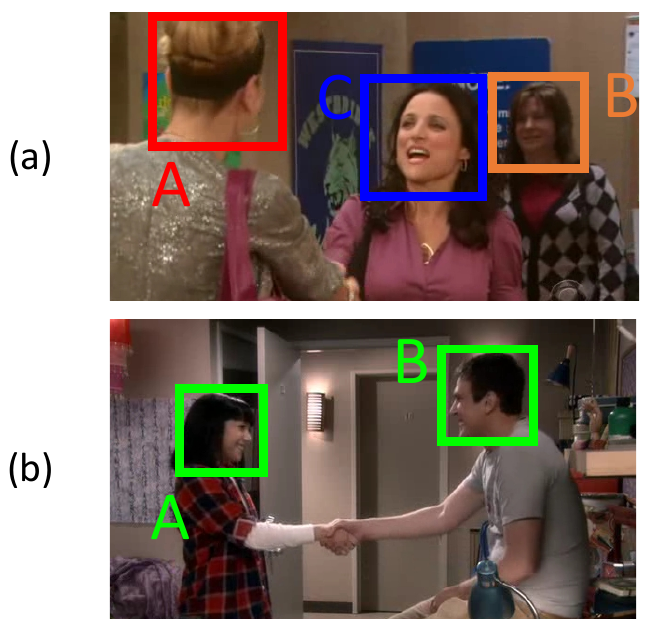}
    \captionsetup{skip=0cm, width=0.8\textwidth}
    \captionof{figure}{\small (a) AB are not LAEO as C is occluding. (b) AB are LAEO.}
    \label{fig:whyhmap}
\end{minipage}

\end{table*}


Given a video clip, we aim to determine if any two humans are \textit{Looking At Each Other} (LAEO). 
To this end, we introduce the \newnet, a three branch \textit{track} network, which takes as input two head tracks and the relative position between the two heads encoded by a head-map, and determines a confidence score on whether the two people are looking at each other or not, and the frames where LAEO occurs. 
The network is applied exhaustively over all pairs of simultaneous head tracks in the video clip.

\newnet consists of three input branches, a fusion block, and a fully-connected layer and is illustrated in Figure~\ref{fig:main}.  Two of the input streams determine the pose of the heads (green branches) and the third represents their relative position and scale (red branch). 
The fusion block combines the embeddings from the three branches and passes them through a fully-connected layer that predicts the LAEO classification (blue layer). 
The network uses spatio-temporal 3D convolutions and can be applied to the head tracks in the video. 
We next describe the components and report their specifications in Table 1 in the supplementary material.

\paragraph{Head-pose branch. }
It consists of two branches, one per person. 
The input of each branch is a tensor of $\mathcal{T}$ RGB frame crops of size $64 \times 64$ pixels, containing a sequence of heads of the same person.
Each branch encodes the head frame crop, taking into account the head pose. 
The architecture of the head-pose branch is inspired by the encoder of the self-supervised  method~\cite{wiles2018bmvc}. It consists of five conv layers, which are followed by a dropout and a flatten ones (green branches in Figure~\ref{fig:main}). 
The output of the flatten layer is L2-normalized before using it for further processing.
Note that the head sequence of each person of the target pair will be encoded by this branch, obtaining two embedding vectors as a result.

\paragraph{Head-map branch. }
This branch embeds the relative position and relative distance to the camera (\ie depth) between two head tracks over time using a head-map. 
In particular, we depict as 2D Gaussians all the heads detected at each frame of the $\mathcal{T}$-frames track (Figure~\ref{fig:headmaps_synpairs}~(a)), whose size is proportional to the head size (\ie detection bounding-box). The different Gaussian sizes encode the relative 3D arrangement (depth) of people in the scene, \ie smaller sizes indicate that people are further from the camera compared to those with bigger size. 
We define a $64 \times 64 \times \mathcal{M}$ map (for the whole $\mathcal{T}$-frames track) that encodes this information\footnote{
Assuming a 0-indexed list, the central frame of a sequence with length $T$ is the $\lfloor T/2 \rfloor$-th one. Specifically, for $T=10$ and $M=1$, the central frame is the one in position 5 (\ie the 6th), whereas for  $M=5$ we use 5 consecutive frames in the central part (taking into account the previous criterion). 
}.  
In addition to the two head tracks, this branch encodes information for other people in the scene. 
Depending on its size and scale, a third person could cut the \textit{gaze ray} between the two side people (Figure~\ref{fig:whyhmap}). 
Including this information helps the \newnet to distinguish such cases. 
This branch consists of a series of four convolutional layers: 
either 2D if we are modeling the relative head position only at the central frame or 3D if we target the whole $\mathcal{T}$-frame head track. 
To obtain the embedding of the head-map we flatten the output of the last conv layer and apply L2-normalization. 

\paragraph{Fusion block.} 
The embedding vectors obtained as the output of the different branches of the network are concatenated and further processed by one fully-connected layer with a dropout layer (blue layer in Figure~\ref{fig:main}). 
Then, a Softmax layer consisting of two hidden units (\ie representing not-LAEO and LAEO classes) follows.

\paragraph{LAEO loss function. }
For training the LAEO predictor, we use the standard binary cross entropy loss: 
\begin{equation}
\label{eq:laeoLoss}
\small{{\mathcal{L}_{\textrm{LAEO}} = - \left( c \cdot \log (\hat{p}_{c}) + (1-c) \cdot \log(1-\hat{p}_{c}) \right),}}
\end{equation}
where $c$ is the ground-truth class ($0$ for not-LAEO, $1$ for LAEO) and $\hat{p}_{c}$ the predicted probability of the pair being class $c$.

\paragraph{Differences between \oldnet~\cite{marin19cvpr} and \newnet. }
\oldnet exploits the temporal information of videos by using as input two head tracks instead of single frames. Nevertheless, the relative position between the two heads (and any interleaving head) is encoded by a \emph{single} frame, \ie one head map. 
We consider this a wasted opportunity, as this single frame may suffer from several issues, such as noise, inconsistency, detection problems, \etc. 
Therefore, we extend the temporal dimension of the head maps and consider multiple consecutive frames instead of single frames. 
This leads to two main architecture changes in \newnet: (a) we consider $\mathcal{M}$-length head-maps, and (b) we decode the information from the temporal sequence of head-maps using a series of 3D conv layers instead of 2D ones from \oldnet~\cite{marin19cvpr} (bottom branch in Figure~\ref{fig:main}). 
Additionally, we change the architecture of the branches that process the head-tracks from a shallower arbitrary-chosen architecture of \oldnet~\cite{marin19cvpr} to the deeper, inspired-by-\cite{wiles2018bmvc} architecture of \newnet. 
\newnet has more parameters and therefore, ability to better generalize and learn better features. In 
Section~\ref{sub:old_vs_new} we present experiments demonstrating the benefit of all changes.

\section{Datasets} 
\label{sec:datasets}

In this section, we describe the LAEO datasets. 
First, we introduce two new datasets: \UCOLAEO and AVA-LAEO, and 
then, two other datasets: AFLW~\cite{koestinger11aflw}, and TVHID~\cite{Patron2010hi5}. 
AFLW is used for pre-training the head-pose branch and for generating synthetic data, while TVHID is used only for testing. 
The newly introduced \UCOLAEO and AVA-LAEO datasets are used both for training and testing \newnet. 
Table~\ref{tab:dbstats} shows an overview of the LAEO datasets. 
The new datasets with their annotations and the code for evaluation are available online at:  \url{http://www.robots.ox.ac.uk/~vgg/research/laeonet/}.

\subsection{The \UCOLAEO dataset }
\label{sub:dat_laeo}
\begin{figure}[t]
\begin{center}
\includegraphics[width=1\linewidth]{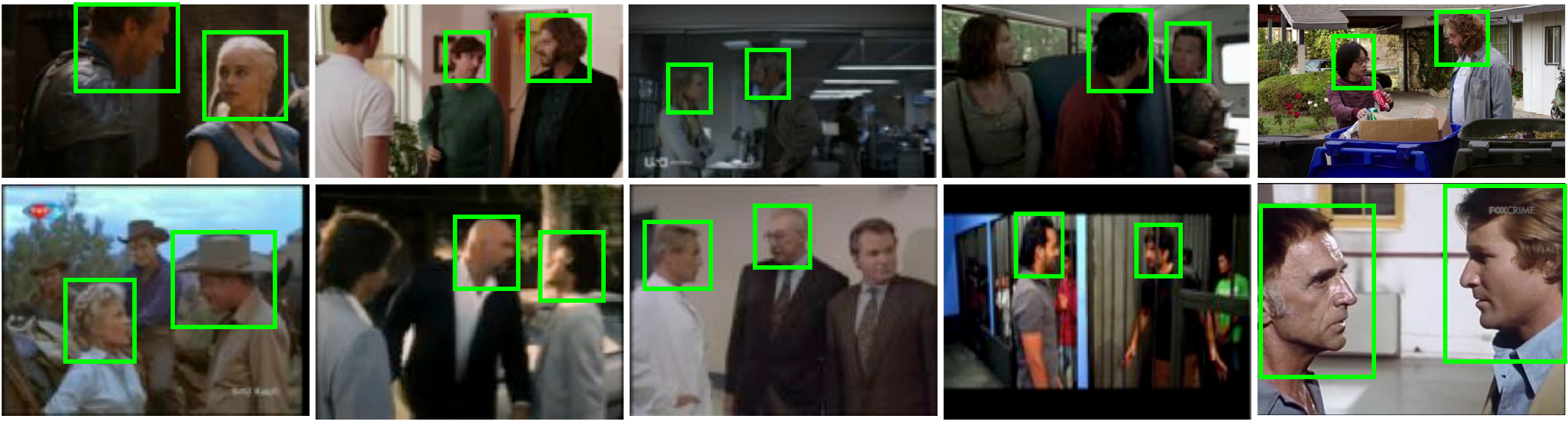}
\end{center}
\beforecaptions
   \caption{\small{\textbf{(top) \UCOLAEO and (bottom) AVA-LAEO datasets.} 
      Example of frames and LAEO head pair annotations included in our new datasets. Different scenarios, people clothing, background clutter and diverse video resolutions, among other factors, make them challenging.
   }}
\label{fig:datasets}
\end{figure}

We use four popular TV shows: `Game of Thrones', `Mr Robot', `Silicon Valley' and `The Walking Dead'. 
From these shows, we collect $129$ ($3$-$12$ seconds long) shots, annotate all the heads in each frame with bounding boxes, and then annotate each head pair as LAEO or not-LAEO (Figure~\ref{fig:datasets}~(top)).

\paragraph{Annotation setup. } 
We annotate all frames both at the frame level, \ie, \textit{does this frame contain any pair of people LAEO?}; and  
at the head level, \ie we annotate all heads in a frame with a bounding-box and all the possible LAEO pairs. 
The visually ambiguous cases are assigned as `ambiguous' and we exclude them from our experiments. 
We split the $100$ LAEO shots into $77$ train, $8$ validation and $15$ test, respectively. This results in $\sim7.5$k training, $\sim1.2$k val and $\sim1.5$k test LAEO pairs (Table~\ref{tab:dbstats}).

\subsection{AVA-LAEO dataset }
\label{sub:dat_ava}
AVA-LAEO consists of movies from the training and validation sets of the `Atomic Visual Actions' dataset (AVA v2.2)~\cite{gu2018ava} dataset. 
The AVA frames are annotated (every one second) with bounding-boxes for $80$ actions, without LAEO annotations; therefore, we enhance the labels of the existing (person) bounding-boxes in a subset of the train and val sets with LAEO annotations. 

\paragraph{Annotation setup. }
From the train and val sets of AVA, we select the frames with more than one person annotated as \textit{`watch (a person)'}, resulting in a total of $40,166$ and $10,631$ frames, respectively. 
We only consider the cases, where both the watcher and the watched person are visible (since the watched person may not be visible in the frame). 
For annotating, we follow the same process as in \UCOLAEO, \ie we annotate each pair of human bounding boxes at the frame level as LAEO, not-LAEO, or ambiguous. 
This results in $\sim19$k LAEO and $\sim118$k not-LAEO pairs for the training set and $\sim5.8$k LAEO and $\sim28$k not-LAEO pairs for the val set (Table~\ref{tab:dbstats}). 
We refer to this subset as AVA-LAEO. 
Figure~\ref{fig:datasets}~(bottom) shows some LAEO pair examples.

\subsection{Additional datasets}

\paragraph{AFLW dataset. } 
\label{subsec:aflw}
We use the `Annotated Facial Landmarks in the Wild' dataset~\cite{koestinger11aflw} to 
(a) pre-train the head-pose branch (first stage, Section~\ref{sub:pretrain-AFLW}), and 
(b) generate synthetic data for training (second stage, Section~\ref{sub:finetune}). 
It contains about $25$k annotated faces in images obtained from FlickR, where each face is annotated with a set of facial landmarks. 
From those landmarks, the head pose (\ie yaw, pitch and roll angles) is estimated. 
To create a sequence of head-crops, we replicate the input image $\mathcal{T}$ times. We keep the two middle replicas unchanged and randomly perturbing the others, \ie small shift, zooming and brightness change.

\paragraph{TVHID-LAEO. } 
\label{subsec:tvhid}
TVHID~\cite{Patron2010hi5} was originally designed for human interaction recognition in videos. 
It contains $300$ video clips with five classes: hand-shake, high-five, hug, kiss and negative. 
We use the LAEO annotations at the shot level from \cite{marin2013ijcv}, which result in $443$ shots with $331$ LAEO and $112$ not-LAEO pairs (Table~\ref{tab:dbstats}). 

\section{Head detection and tracking} 
\label{sub:d_t}
Unlike most methods that rely on faces, \newnet requires {\em head} tracks as input. 
Here, we train the Single Shot Multibox Detector detector~\cite{liu2016ssd} from scratch and obtain head detections. 
Then, we group them into tracks using the linking algorithm from~\cite{kalogeiton2017action} (see Section 2 
in the supplementary material).
\section{Training \newnet}
\label{sec:training}

We describe here our two-stage training procedure. 
The first stage involves only the head-pose branches (Section~\ref{sub:head-pose}).  
We consider three initialization options for these branches: 
(i) fully-supervised pre-training with annotated head-pose data (Section~\ref{sub:pretrain-AFLW}), 
(ii) self-supervised pre-training using Facial Attributes-Net \cite{wiles2018bmvc} (Section~\ref{sub:pretrain-ss}), 
or (iii) completely random initialization, \ie no pre-training (Section~\ref{sub:pretrain-random}). 
In the second stage, we train \newnet from scratch, \ie head-map and upper layers (Section~\ref{sub:finetune}).

\subsection{Head-pose branches}
\label{sub:head-pose}
In general, humans can infer \textit{where} a person is looking just based on the head pose, without even seeing the eyes~\cite{langton2004influence}. 
This shows that important information is encoded in the head orientation. 
In the literature, several works model the head orientation \cite{ruiz2018hpose} or the eye gaze~\cite{recasens2015nips}.  
Note that using the actual eye gazing is not always an option, even with multiple-frames as input, as there is no guarantee that the eyes are fully visible, \ie due to image resolution, or self occlusions. 
Therefore, in this work we model gaze just based on head orientation. 
In particular, we either
(i)~pre-train a model that learns the head orientation using the head angles (Section~\ref{sub:pretrain-AFLW}), or
(ii)~use the self-supervised Facial Attributes-Net that models the head orientation implicitly (Section~\ref{sub:pretrain-ss}), or  
(iii)~use a random initialization for the \newnet that manages to learn the head-pose orientation (Section~\ref{sub:pretrain-random}).

\subsubsection{Fully-supervised pre-training}
\label{sub:pretrain-AFLW}

We model head orientation with three angles (in order of decreasing information): (a)~yaw, \ie looking right, left, (b)~pitch, \ie looking up, down, and (c)~roll, \ie in-plane rotation. 
We use this modelling to pre-train the head-pose branches. 

\paragraph{Loss function of head-pose pre-training. }
Let $(\alpha, \beta, \gamma)$ be the yaw, pitch and roll angles of a head, respectively. 
We define one loss for estimating each pose angle: $\mathcal{L}_{\alpha}$, $\mathcal{L}_{\beta}$, $\mathcal{L}_{\gamma}$ and model them with the $L1$-smooth loss~\cite{ren2015fastrcnn}. 

Given that the yaw angle is the dominant one, in addition to these losses, we include a term that penalizes an incorrect estimation of the sign of the yaw angle, \ie, failing to decide if the person is looking left or right ($\mathcal{L}_{s}$). It is defined as: 
\begin{equation}
\mathcal{L}_{s} = \max(0, - \mathrm{sign}(\alpha) \cdot \mathrm{sign}(\hat{\alpha}) ) , 
\end{equation}
where $\mathrm{sign}(\alpha)$ is the sign function (\ie returns $+1$ for positive inputs, $-1$ for negative inputs, and $0$ if the input is $0$) applied to the yaw angle; and, $\hat{\alpha}$ is the ground-truth angle. 
In practise, as the gradient for the sign function is always 0, it is implemented by using $\mathrm{tanh}(\cdot)$ (hyperbolic tangent).

Therefore, the loss function $\mathcal{L}_h$ for training the head-pose branch for LAEO purposes is given by:
\begin{equation} \label{eq:headloss}
    \mathcal{L}_h = w_{\alpha} \cdot \mathcal{L}_{\alpha} + w_{\beta} \cdot \mathcal{L}_{\beta} + w_{\gamma} \cdot \mathcal{L}_{\gamma} + w_s \cdot \mathcal{L}_s, 
\end{equation}
where $w_x$ are positive weights chosen through cross-validation at training.
In our experiments, we use: $w_{\alpha}=0.6$, $w_{\beta}=0.3$, $w_{\gamma}=0.1$, $w_s=0.1$, as $w_{\alpha}$ is the dominant one. 
Note that the weights do not necessarily add to $1$. 
Please refer to Section 3 in the supplementary material for ablations about the two losses.

\subsubsection{Self-supervised pre-training}
\label{sub:pretrain-ss}
The goal is to use a (self-supervised) network that learns head-pose orientation without being explicitly trained on it. 
To this end, we use a modified version of the self-supervised Facial Attributes-Net from~\cite{wiles2018bmvc}. 
The Facial Attributes-Net uses a single frame, whereas we are interested in $\mathcal{T}$ input video frames. 
Therefore, we inflate the filters from the Facial Attributes-Net for $\mathcal{T}$ consecutive frames, by replicating their weights. 
Moreover, we change the input size of the Facial Attributes-Net from $256\times256$ to the input of \newnet, \ie $64\times64$.

\subsubsection{Random initialization}
\label{sub:pretrain-random}
For reference, we also initialize the \newnet with random values for the weights. 
Albeit randomly initialized, \newnet manages to learn head pose and orientation implicitly by solving the LAEO task alone (Section~\ref{sub:pre-training}).

\subsection{Training the \newnet} 
\label{sub:finetune}

We train \newnet with both real and synthetic data. 
We use data augmentation techniques, such as image perturbations, translations, brightness changes, zoom changes, \etc. 
For the first $N=2$ epochs, we use only synthetic data, and then we alternate between real and synthetic data. 
To improve the performance of the model, we use hard negative mining. 
We deploy the curriculum learning strategy of \cite{Nagrani18c}, which modulates the difficulty of the hard negatives incorporated into training. 
In our experiments, the value of the negative difficulty parameter~\cite{Nagrani18c} 
is increased after $2$ epochs, allowing more difficult samples as its value increases.

\begin{figure}[t!]
\centerline{%
\begin{tabular}{c@{}c@{}}
\includegraphics[width=0.52\linewidth]{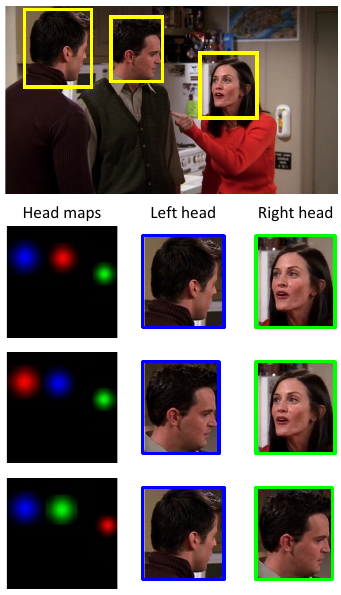}& 
\includegraphics[width=0.46\linewidth]{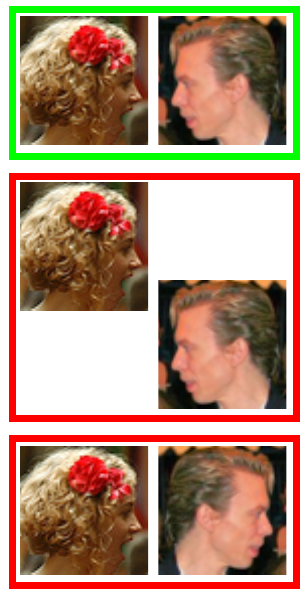} \\
\small{(a)} & \small{(b)} \\
\end{tabular}}
\beforecaptions
\caption{\small{\textbf{(a)~Head-maps and (b)~augmentation of LAEO samples.} 
(a)~We analyse all head pairs with a color coding: \textit{blue} for the left, \textit{green} for the right and \textit{red} for the remaining heads, such as middle, \ie not considered for evaluation.  
(b)~We generate synthetic LAEO negative training data (red boxes) from positive pairs (green box), based on the orientation or the relative position of the heads. 
}}
\label{fig:headmaps_synpairs}
\end{figure} 

\paragraph{Synthetic data. }
For generating synthetic data we use images with head-pose information. 
To generate positive samples, we select pairs of heads whose angles are compatible with LAEO and, at the same time, they generate consistent geometrical information. 
To generate negative samples, we either
(i)~change the geometry of the pair, \ie making LAEO not possible any more, \eg by mirroring just one of the two heads, or 
(ii)~select pairs whose pose are incompatible with LAEO, \eg both looking at the same direction. 
Figure~\ref{fig:headmaps_synpairs}~(b) shows some artificially generated pairs.

\begin{figure*}[ht]
\begin{center}
\includegraphics[width=1\linewidth]{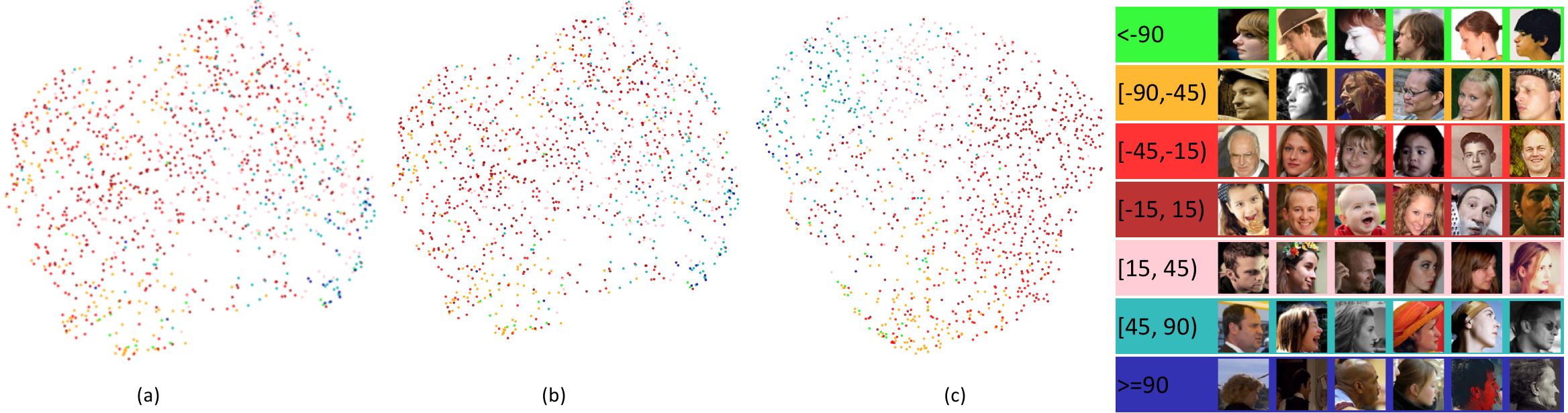}
\end{center}
 \caption{\small{\textbf{Head embeddings automatically learnt during LAEO training with full random initialization} after (a) one, (b) ten, (c) twenty epochs. \newnet is paying attention to head orientation to solve the task, and therefore the head pose is learnt implicitly by the LAEO task alone (implicit supervision). The set of discretized angles obtained from AFLW dataset are shown in the legend. Note that the more the network learns the LAEO task, the more the clusters become separate (\eg yellow points that correspond to [-90,45) angles). On the right, we illustrate head crops for the discretized set of angles after the training has finished. 
 (\textit{Best viewed in digital format.})
 }}
 \label{fig:headembed}
\end{figure*}

\section{Evaluation and scoring methodology}
\label{sec:metrics}

\paragraph{LAEO-classification AP} 
is the metric we use to evaluate the LAEO predictions. 
Similar to object detection, a detection is correct if its intersection-over-union overlap (IoU) with the ground-truth box is $>0.5$ \cite{voc}. 
A detected pair is correct if both heads are correctly localized and its label (LAEO, not-LAEO) is correct. 
The performance is Average Precision (AP) computed as the area under the Precision-Recall (PR) curve. 
Depending on the available ground-truth annotations, we measure AP at frame level, considering each pair as an independent sample, or at shot-level, if more detailed annotations are not available. 
Frame level is used for \textit{\UCOLAEO} and \textit{AVA-LAEO} and, following previous work~\cite{marin2013ijcv,masse2018pami}, shot level for \textit{TVHID}.

\paragraph{Scoring methodology. }
Given that the level of (ground truth) annotation differs between the three datasets, we describe how we use the \newnet outputs to obtain the final scores, either at the shot or at the frame level. 
We test \newnet on pairs of head-tracks (of length $\mathcal{T}=10$), obtain one LAEO score for each track-pair, and assign the LAEO score to the head-pair in the middle frame. 
The scoring process for each dataset is as follows:

\begin{enumerate}[label=(\roman*)]
\item \textit{\UCOLAEO}:
Since the bounding boxes for the heads are available for each frame, the \newnet is applied directly to these head tracks (no detections are used). 
To account for the $T/2$ frames at the beginning (resp.\ end) of a track, we propagate the score from the middle frame. 
\item \textit{AVA-LAEO:}
We run the head tracker and apply \newnet on these tracks. 
AVA-LAEO contains pair annotations for \textit{human} bounding-boxes (instead of heads); hence,
we compare each head pair against the ground-truth human pairs using intersection over head area (instead of IoU).
\item \textit{TVHID:}
We run the head tracker and apply \newnet on the tracks. We compute a LAEO score as the max of 
smoothed scores in a shot; 
the smoothed score is the average of a moving temporal window (of length five) along the track.
\end{enumerate}

\section{Experimental results}
\label{sec:expers}

In this section, we experimentally evaluate the effectiveness of \newnet for determining people LAEO. 
Note that the model is trained either on \UCOLAEO or on AVA-LAEO. 
Here, we study the impact of all training and architecture choices.

First, we examine the importance of 
the head-map branch, 
the length $\mathcal{T}$ of the head-tracks, and  the length of $\mathcal{M}$ of the head-map (Sections~\ref{sub:head-map}-\ref{sub:Mlength}). 
Then, we assess the importance of the different pre-training schemes (Section~\ref{sub:pre-training}). 
In Section~\ref{sub:res_newdatasets} we examine the performance of \newnet on the two new test datasets, \UCOLAEO, and AVA-LEO. 
After, we analyse different domains by performing a cross dataset evaluation (Section~\ref{sub:cross}). 
In Section~\ref{sub:old_vs_new} we provide experimental results between \oldnet and \newnet, and n Section~\ref{sub:discussion} we provide a summary of our findings. 
Finally, in Section~\ref{sub:restvhid}, we compare \newnet to \oldnet~\cite{marin19cvpr} and to other state-of-the-art methods on the \UCOLAEO, AVA-LAEO, and TVHID-LAEO datasets.

\paragraph{Implementation details.}
\newnet is implemented with Keras~\cite{chollet2015keras} using TensorFlow as backend. All implementation details can be found in Section 1.2 in the supplementary material.

\subsection{Importance of the head-map} 
\label{sub:head-map}
We evaluate \newnet with and without the head-map branch (Table~\ref{tab:ablation2}). 
Adding it improves the performance (from $80.3\%$ to $\UCOLAEOscoreTrUCOSSMA\%$ for $\mathcal{T}{=}10$), as it learns the spatial relation between heads.

\paragraph{Comparison with the geometry branch baseline.} 
To assess the quality of the head-maps branch, we consider a baseline: 
the \textit{geometrical information branch}, where the relative position of two heads over time is encoded by their geometry. It embeds the relative position between two head tracks over time (relative to a $(1,1)$ normalized reference system), and the relative scale of the head tracks. The input is a tuple $(dx, dy, s_{r})$, where $dx$ and $dy$ are the $x$ and $y$ components of the vector that goes from the left head $L$ to the right one $R$, and $s_{r} = s_{L}/s_{R}$, is the ratio between the scale of the left and right heads. 
The consists of two fc layers with 64 and 16 hidden units and it outputs a vector of 16 dimensions encoding the geometrical relation between the two target heads. 
\newnet with the geometry branch results in $1\%$ less classification AP than with the head-pose branch. 
This is expected; even though both branches encode the same information (\ie relative position of the two heads), the head-maps branch provides a richer representation of the scene, as it encodes information for all existing heads and, therefore, results in better AP. Note, using both the head-map \textit{and} the geometry branches (in addition to the head-pose branches) does not lead to any further improvement, as the combination of these two branches just increases the number of parameters without providing additional information. 
Thus, we conclude that \newnet is the most effective architecture in terms of AP performance.

\subsection{Temporal window $\mathcal{T}$} 
\label{sub:Klength}
To assess the importance of the temporal window using $\mathcal{T}$ frames compared to using a single frame, we vary $\mathcal{T}$ and train and evaluate \newnet with $\mathcal{T}=1,5,10$. 
Table~\ref{tab:ablation2} shows that there is an improvement in AP performance of $1.5\%$ when $\mathcal{T}$ increases from only 1 to 5 frames, and a significant improvement of $2.9\%$ when $\mathcal{T}$ increases from 1 to 10 frames
(we found no improvement for $\mathcal{T} > 10$). 
In the remainder of this work, we use $\mathcal{T}=10$ frames.
%

\subsection{Length of Head-map $\mathcal{M}$}
\label{sub:Mlength}

To assess the importance of the length of the head-map $\mathcal{M}$ compared to using a single-frame, we vary $\mathcal{M}$ and train and evaluate \newnet with $\mathcal{M}=1,5,10$ with various pre-training schemes. 
Table~\ref{tab:ablation45} shows the results for \UCOLAEO and AVA-LAEO. 
We observe that there is a significant performance improvement of approximately $5\%$ when increasing the length of the head-map from 1 to 10 for all cases. 
Therefore, for the remainder of this work, we use $\mathcal{M}=10$.

\begin{table*}
\begin{minipage}[b]{0.25\linewidth}
\centering
\footnotesize{{
\begin{tabular}{crrr}
\toprule
$\mathcal{M}$= & \multicolumn{3}{c}{$\mathcal{T}$=} \\ 
1 & 1 (2D) & 5 & 10 \\
\midrule
- &  77.5 &  78.2 & 80.3 \\ 
\checkmark & 78.6 & 80.1 & 
\textbf{\UCOLAEOscoreTrUCOSSMA} \\ 
\bottomrule
\end{tabular}}}
\vspace{10mm}
\caption{\small{\textbf{Head-map and temporal window $\mathcal{T}$}. \%AP when training and testing on \UCOLAEO w and w/o the head-map for $\mathcal{T}$=1,5,10. 
}}
\label{tab:ablation2}
\end{minipage}
\hspace{0.15cm}
\begin{minipage}[b]{0.38\linewidth}
\centering
\small{{
\resizebox{\linewidth}{!}{
\begin{tabular}{lrrrrrr}
\toprule
pre-train / & \multicolumn{3}{c}{\UCOLAEO} & \multicolumn{3}{c}{AVA-LAEO}  \\
 $\mathcal{M}=$  & 1 & 5 & 10 & 1 & 5 & 10  \\
\midrule
random & 78.4 & 78.5 & 75.3 & 58.3 & 64.7 & 67.6\\
AFLW & 80.2 & 78.7 & 83.7 & 59.3 & 67.0 & 68.6 \\
self-superv. & \UCOLAEOscoreTrUCOSSMA & \UCOLAEOscoreTrUCOSSMB & \textbf{\UCOLAEOscoreTrUCOSSMC} & \AVALAEOscoreTrAVASSMA &  \AVALAEOscoreTrAVASSMB & \textbf{\AVALAEOscoreTrAVASSMC} \\
\bottomrule
\multicolumn{4}{l}{
\small{${}^{\dagger}$ head-track $\mathcal{T}$=10}
}
\end{tabular}}}}
\vspace{4.2mm}
\caption{\small{\textbf{Head-map length $\mathcal{M}$ and pre-training schemes}. We report \%AP when training and testing on \UCOLAEO \major{and AVA-LAEO} for head-maps of $\mathcal{M}$=1,5,10 with different pre-training schemes. 
}}
\label{tab:ablation45}
\end{minipage}
\hspace{0.15cm}
\begin{minipage}[b]{0.36\linewidth}
\centering
\small{{
\resizebox{\linewidth}{!}{
\begin{tabular}{llcrr}
\toprule
\multirow{2}{*}{net} & \multirow{2}{*}{pre-train}  & \multirow{2}{*}{$\mathcal{M}$=}  & UCO- & AVA-  \\
& & & LAEO & LAEO \\
\midrule

\oldnet~\cite{marin19cvpr} & AFLW & 1 & 79.5 & 50.6  \\
\newnet & AFLW & 1 & 80.2  & 59.3   \\ 
\newnet & self-superv. & 1 & \UCOLAEOscoreTrUCOSSMA  & \AVALAEOscoreTrAVASSMA  \\
\newnet & AFLW & 10 & 83.7  & 68.6 \\

\newnet & self-superv. & 10 & \textbf{\UCOLAEOscoreTrUCOSSMC}  &  \textbf{\AVALAEOscoreTrAVASSMC}  \\
\bottomrule
\multicolumn{4}{l}{
\small{${}^{\dagger}$ head-track $\mathcal{T}$=10}
}
\end{tabular}}}}
\caption{\small{\textbf{Comparison between \oldnet and \newnet}. We report \%AP when training and testing on \UCOLAEO and AVA-LAEO for $\mathcal{M}$=1,10 with different pre-training schemes.
}}
\label{tab:v0vsv1}
\end{minipage}
\vspace{1.5mm}
\par
\centering
\small
\begin{tabular}{lccccccc}
\toprule
train on    & \UCOLAEO & AVA-LAEO & \UCOLAEO & AVA-LAEO &  \UCOLAEO & AVA-LAEO & TVHID \\
test on & \multicolumn{2}{c}{\UCOLAEO} & \multicolumn{2}{c}{AVA-LAEO} & \multicolumn{3}{c}{TVHID} \\ 
\midrule
baseline (chance level) &  \multicolumn{2}{c}{\UCOchance} & \multicolumn{2}{c}{\AVAchance} & \multicolumn{3}{c}{--} \\ 
\cite{marin2013ijcv} (Fully auto+HB)  &  --  & -- & -- & -- & -- & -- & 87.6 \\ 
\cite{masse2018pami} (Fine head orientation)  & -- & -- & -- & -- &  -- & -- & 89.0 \\ 
\cite{marin19cvpr} \oldnet(pre-trained)  &\UCOLAEOscore  & \UCOLAEOscoreTrAVA & \AVALAEOscore & \AVALAEOscoreTrAVA & 91.8 & \TVHIDscoreTrAVA & -- \\
\newnet (self-supervised) &  \textbf{\UCOLAEOscoreTrUCOSSMC}  & \UCOLAEOscoreTrAVAFtUCO & \AVALAEOscoreTrUCOFtAVA & \textbf{\AVALAEOscoreTrAVASSMC} & \textbf{\TVHIDscoreBestTrUCO}($\mathcal{M}$=1) & \TVHIDscoreBestTrAVA & -- \\
\bottomrule
\end{tabular}
\vspace{1               mm}
\caption{\small{\textbf{LAEO results on \UCOLAEO, AVA-LAEO and TVHID.} We report \%AP at the pair@frame level for \UCOLAEO and AVA-LAEO and, similar to other works, at the shot level for TVHID. 
}}
\vspace{-0.3cm}
\label{tab:results}
\end{table*}

\subsection{Pre-training schemes}
\label{sub:pre-training}
We examine the three different settings for pre-training \newnet: fully supervised, where we pre-train using ground-truth labels for head orientations (Section~\ref{sub:pretrain-AFLW}); self-supervised, where we employ a video model learnt to solve another task (Section~\ref{sub:pretrain-ss}); and implicit supervision, where we use random initialization (Section~\ref{sub:pretrain-random}). 
Table~\ref{tab:ablation45} reports the \%AP results.

For low values of $\mathcal{M}=1$, the random initialization performs similarly to the other models. 
This shows that for $\mathcal{M}=1$ there is sufficient data to train the network for the LAEO task; however, for higher values of $M$ there is not enough training data. 
It is, therefore, interesting to investigate the learnt properties of the network when $\mathcal{M}=1$. 
To this end, we evaluate \newnet trained with random intialization on AFLW, which has ground truth labels for the head orientations. 
We project the predicted head-embeddings on a 2D space using the Uniform Manifold Approximation and Projection for Dimension Reduction ~\cite{umap} and illustrate it in Figure~\ref{fig:headembed}. 
\newnet groups the heads based on their orientation (we depict discretized angles). 
Specifically, we illustrate the  head embeddings after one, ten and twenty epochs of training and observe that the more \newnet is trained, the more separate the head clusters become. 
Thus, we conclude that to solve an explicit task, \ie people LAEO, \newnet learns an additional task, \ie estimating head pose (implicit supervision).

For longer temporal head-maps, \eg $\mathcal{M}{=}10$, the self-supervised model outperforms the other ones by a small margin ($1$-$3\%$). 
This is interesting as one might expect the fully supervised one to prevail.  
This is probably due to the size and variety of the training data: the self-supervised model has been trained on a larger dataset~\cite{voxceleb2} and with greater pose variation than the one in AFLW. 
Overall, the self-supervised pre-training outperforms the rest; hence, in the remainder of this work we use this. 

\vspace{-0.2cm}
\paragraph{Relation to gaze direction.}
An alternative pre-training scheme would be to use gaze direction models as initialization for \newnet. For instance, the head-branch could be initialized by the one from~\cite{recasens2015nips}, as both encode information about the head pose and orientation. 
Similarly, \newnet could be used as initialization for gaze direction models~\cite{recasens2015nips,recasens2017iccv}. 
Moreover, \newnet could be adapted to infer person-wise VFoA, for instance by replacing one head-track branch by a saliency predictor~\cite{recasens2015nips} or a transformation pathway~\cite{recasens2017iccv}. 
A possible extension would be to add saliency prediction in \newnet as additional task for joint training with the LAEO.
Another line of work would be to combine the self-supervised LAEO pre-training with 3D gaze estimation~\cite{doosti2020mgaze3d} to scale-up gaze estimation or gaze following.

\subsection{Results on \UCOLAEO and AVA-LAEO}
\label{sub:res_newdatasets}
Table~\ref{tab:ablation45} reports the results when evaluating  \newnet on \UCOLAEO and AVA-LAEO. 
The performance is $86.7\%$ and $68.7\%$ when training and testing on \UCOLAEO, and AVA-LAEO, respectively. 
These results reveal that there exists a significant gap in the performance between the two datasets. 
This is due to the different nature of AVA-LAEO compared to other datasets: 
(1) head annotations are not provided (just human bounding-boxes every 1 second); 
(2) it contains challenging visual concepts, such as (a) low resolution movies, (b) many people in a scene, (c) blurry, small heads, and (d) particular clothing styles, \eg several people wearing hats (western, Egyptian's, turbans, \etc). 
Despite these difficulties, \newnet achieves AP=$68.7\%$.

Moreover, to assess the difficulty of these
datasets and the effectiveness of \newnet, we compare it to the chance level classification. 
\newnet outperforms chance level by a large margin: $\times 2$ for UCO and $\times 4$ for AVA (Table~\ref{tab:results}).

\begin{figure}[t]
\begin{center}
\includegraphics[width=\linewidth]{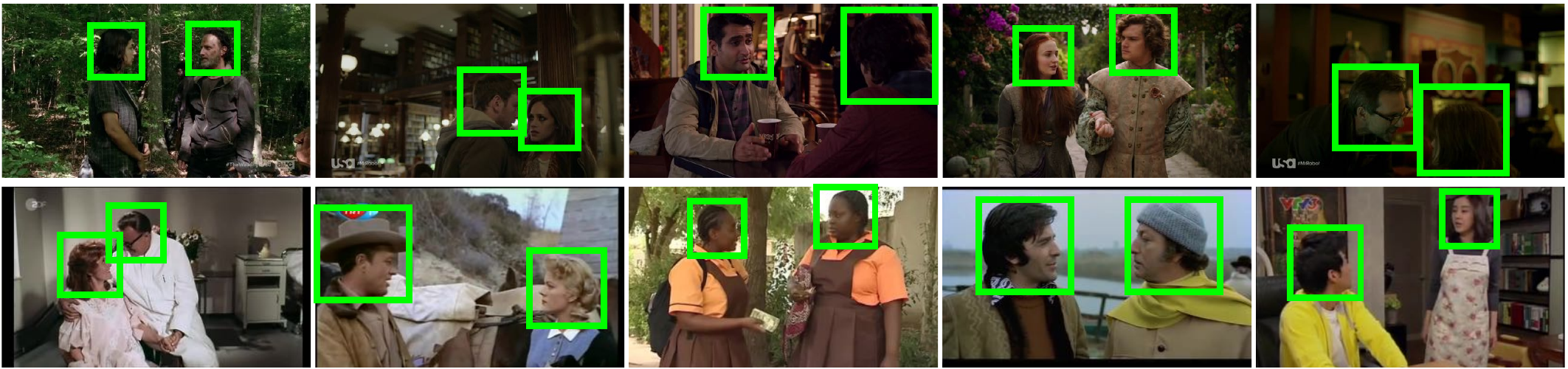}
\end{center}
\beforecaptions
   \caption{\small{\textbf{\newnet results on  \UCOLAEO (top) and AVA-LAEO (bottom).} 
    For different scenarios, backgrounds, head poses \etc, in most cases  \newnet successfully determines if two people are LAEO (green boxes). 
    }}
\label{fig:res_datasets}
\end{figure}

When applying \newnet on UCO and AVA we obtain the results of Figure~\ref{fig:res_datasets}, where we display some of the highest ranked pairs of people LAEO. 
We observe that \newnet leverages the head orientations and their temporal consistency and accurately determines the frames where people are LAEO. 

We hope that \newnet with these two datasets will provide solid baselines and help future research on this area. 

\paragraph{Impact of the detection and tracking errors on the AP.}
\newnet ($T{=}10$,  $M{=}10$) achieves an AP=68.7\% when evaluated on all annotated pairs of AVA-LAEO.
In contrast, if we compute the LAEO classification accuracy only on the subset of \emph{detected pairs}, the AP increases up to 79.8\%.

\vspace{-3mm}
\subsection{Cross-dataset evaluation}
\label{sub:cross}
Here, we aim at examining the generalization of \newnet accross different domains. 
To this end, we examine the performance when initializing the weights with one dataset and fine-tuning it (and testing) on another dataset, \ie pre-training on UCO (and fine-tune on AVA) leads to 67.0\% AP, whereas pre-training on AVA (and fine-tune on UCO) to 84.5\%. 
Interestingly, we observe that this cross-dataset scheme performs very good, resulting in classification performances similar to the ones with no change in domain: 
for UCO there is a drop of only 2.2\% ($84.5\%$ vs $86.7\%$), and for AVA the drop is 1.7\% ($67.0\%$ vs $68.7\%$) Table~\ref{tab:ablation45}. 
These results show that the domain shift~\cite{torralba11cvpr} definitely affects the performance, and that for solving the LAEO task, the pre-training is less important than the actual data for fine-tuning.

\subsection{Comparison between \oldnet and \newnet}
\label{sub:old_vs_new}
We examine the differences in performance between \oldnet and \newnet for the \textit{same} setting (paragraph with differences in Section~\ref{sec:model}).
Table~\ref{tab:v0vsv1} reports the \% AP results when training and testing the two networks on \UCOLAEO and AVA-LAEO.

\textit{Single frame head-map: }
\oldnet results in AP = 79.5\% for UCO and 50.6\% for AVA, whereas for the same setting replacing the head-track architecture of \oldnet with the new one results in AP = 80.2\% for UCO and 59.3\% for AVA, \ie absolute improvements of 0.7\% for UCO and 8.7\% for AVA. 
These suggest that the new architecture helps determining the mutual gaze between people; this is especially demonstrated by the big boost on AVA, 
thus suggesting that the new model handles difficult scenes and scenarios better than the old one, given the more challenging nature of AVA.  
Additionally, using the self-supervised pre-training for \newnet leads to AP = 81.5\% for UCO and 59.8\% for AVA, \ie absolute improvements of 2\% for UCO and 9.2\% for AVA wrt~\cite{marin19cvpr}, showing that the proposed self-supervised pre-training of \newnet leads to greater performance improvements than the AFLW pre-training.

\textit{Multi-frame head-map: }
Increasing the head-map length from one to $\mathcal{M}=10$ frames leads to AP=83.7\% for \UCOLAEO and 68.6\% for AVA-LAEO when using the AFLW pre-training, and AP=86.7\% for \UCOLAEO and 68.7\% for AVA-LAEO when using the self-supervised pre-training. 
The improvements compared to \oldnet are between 4-7\% for \UCOLAEO and around 18\% for AVA-LAEO. 
This clearly indicates that using multiple frames for the head-map boosts the LAEO performance, as the network is better able to capture the temporal aspect of moving heads, thus reducing the missed detections and the false positives.

\subsection{Summary}
\label{sub:discussion}
The findings of \newnet can be summarized as follows: 
\begin{enumerate}[label=(\roman*)]
\item the head-map branch is the most suitable architecture for the task we examine (Table~\ref{tab:ablation2}), 
\item exploiting the temporal dimension by using $\mathcal{T}$-frame long head-tracks and $\mathcal{M}$-frame long head-maps boosts the performance (Tables~\ref{tab:ablation2}-\ref{tab:ablation45}). 
\item for low values of the head-map length ($\mathcal{M}=1$) pre-training is not necessary for solving the LAEO task; nevertheless, for larger values of $\mathcal{M}$ there is a significant benefit in pre-training, as the model benefits from more data (Table~\ref{tab:ablation45}). 
\item solving the LAEO task \textit{alone} (without any pre-training) results in learning head pose and orientations (implicit supervised setting, Figure~\ref{fig:headembed}).
\item AVA-LAEO is more challenging than \UCOLAEO due to its different nature.
\end{enumerate}

\begin{figure}[t]
\begin{center}
\includegraphics[width=1\linewidth]{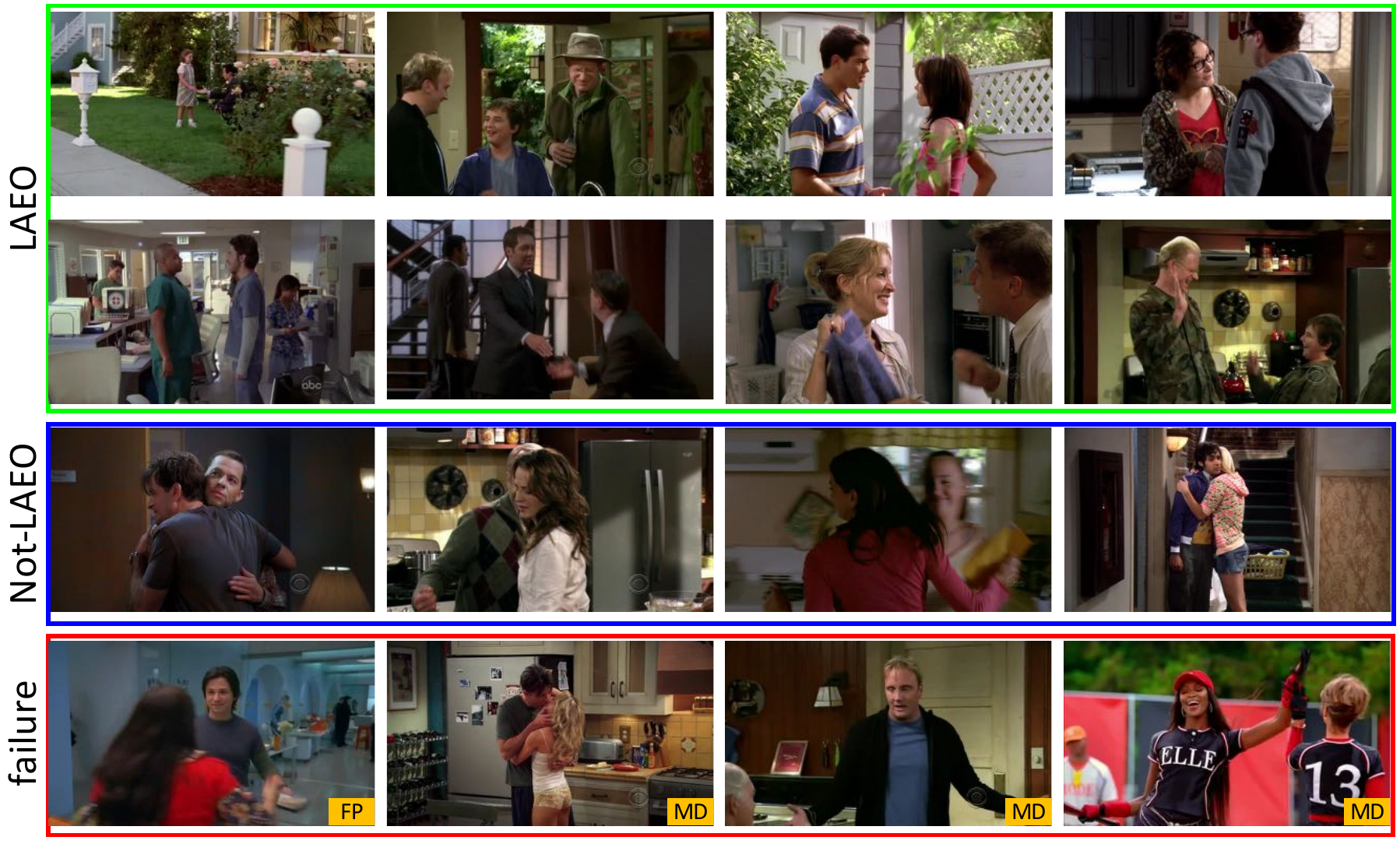}
\end{center}
\beforecaptions
   \caption{\small{\textbf{\newnet results on TVHID.}  
   (top three rows) correct LAEO results when the ground truth is LAEO (green) and not-LAEO (blue).  
   \newnet successfully detects people LAEO in several situations (illuminations, scales, clutter). 
   (last row) failure cases for false positive LAEO detections (first example) and missed detections (three last examples). 
    Most failures are missing people LAEO in ambiguous scenes; \eg in the last red frame the characters are LAEO, even though the character on the left has closed eyes.
   }}
\label{fig:res_tvhid}
\end{figure}

\begin{figure*}[t!]
\begin{center}
\includegraphics[width=1\linewidth]{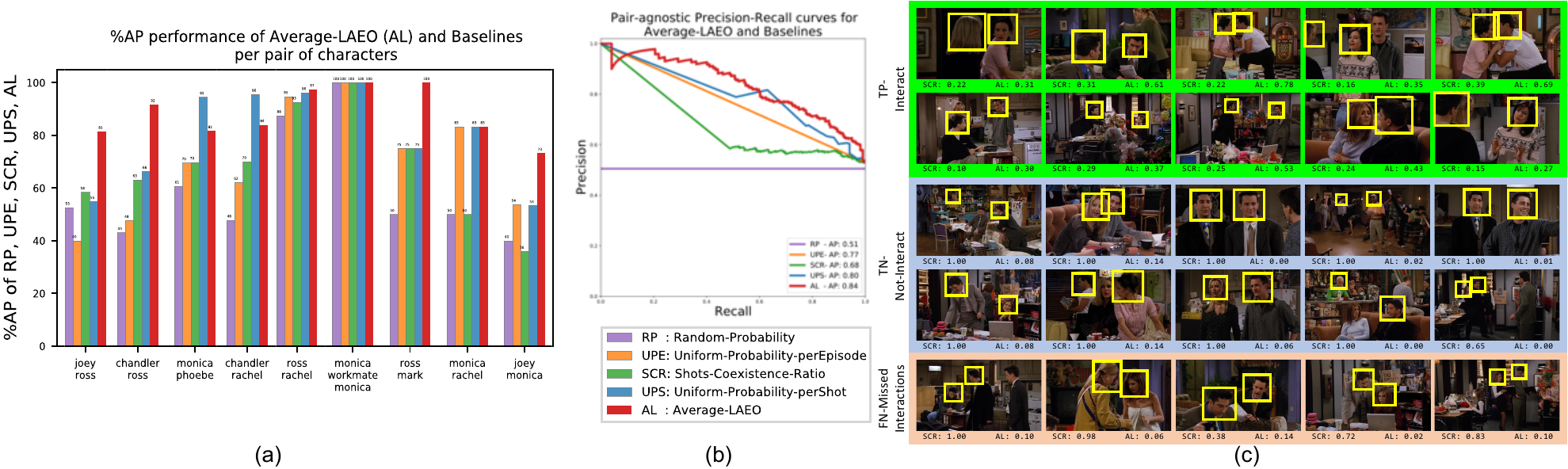}
\end{center}
\caption{
\small{
\textbf{Interaction prediction with the \textit{Average-LAEO} vs various Baselines on Friends.}  
 In addition to the Average-LAEO score (AL), we display four baselines: Random Probability (PR), Uniform Probability per Episode (UPE), Shots-Coexistence-Ratio (SCR), and Uniform Probability per Shot (UPS). 
 (a)~AP performance for AL and various baselines for each pair; (more pairs in Section 4 in supplementary material)
 (b)~Pair-agnostic precision-recall curves. Some patterns are clear: `Ross and Rachel' or `Monica and her workmate' interact with each other almost continuously when they coexist; 
 albeit their low frequency of co-existence, `Joey and Ross', `Joey and Monica' or `Ross and Mark' interact significantly when they co-exist as captured mainly by AL (red). 
 (c)~Examples of AL and SCR. We compute the AL of each pair and display some examples: 
 true positives (TP), when we correctly predict a pair of characters as interacting (green color); 
 true negatives (TN), when we correctly predict a pair of characters as not-interacting (blue color);
 false negatives (FN), when we miss pairs that interact (orange color). 
 Note than in all examples the SCR results are reversed (see SCR scores):
 \ie the green rows are wrongly predicted as not-interacting; 
 the blue rows are wrongly predicted as interacting; 
 the orange row is correctly predicted as interacting. 
 As expected, we observe that the AL fails to determine interactions, where the people are not LAEO. In most cases, however, either in real life or in TV-shows a human interaction typically involves gazing; hence, the AL is suitable for automatically capturing pairs of characters that interact.
 (\textit{Best viewed in digital format.})
 }}
 \aftercaptions
 \label{fig:rankfriends2}
\end{figure*}

\subsection{Results on TVHID-LAEO}
\label{sub:restvhid}

We compare \newnet to the state of the art on TVHID~\cite{Patron2010hi5}, \ie the only video dataset with LAEO annotations (Section~\ref{subsec:tvhid}). 
As in \cite{marin2013ijcv}, we use average AP over the two test sets  (Table~\ref{tab:results}). 
\newnet trained on \UCOLAEO and AVA-LAEO achieves AP$=92.3\%$ and AP=$87.4\%$, respectively. 
Notably, training on \UCOLAEO outperforms training on AVA-LAEO when tested on TVHID.  
This is due to the fact that the domain of TVHID is closer to the one of \UCOLAEO than to AVA-LAEO, given that \UCOLAEO and TVHID consist of TV shows, whereas AVA-LAEO contains movies. 
Despite the domain differences, \newnet trained on AVA-LAEO achieves comparable results to the state of the art. 
Finally, we observe that the model trained on \UCOLAEO outperforms all other methods by a large margin ($1-3\%$).

When we apply \newnet on TVHID and obtain the results shown in Figure~\ref{fig:res_tvhid}. 
Our model successfully detects people LAEO in several situations and scenarios, such as different illuminations, scales, cluttered background. 
By examining the remaining~$8\%$ error, we note that in most cases, the ground truth label is ambiguous, \eg last two red frames in Figure~\ref{fig:res_tvhid}.

\section{Social network \& Interaction prediction}
\label{sec:friends}

One principal way of signaling an interest in social interaction is the willingness of people to LAEO~\cite{goffman2008public,loeb1972mutual}. 
The duration and frequency of eye contact reflects the power relationships, the attraction or the antagonism between people~\cite{abele1986gaze}. 

We present two applications of  \newnet in analysing social interactions in TV material. First,  at the shot level, we show that LAEO is an indicator of whether two characters are {\em interacting} (see below).
Second,
at the episode level, we show
that LAEO is an indicator of the extent of social interactions between two characters, we term this \textit{friend-ness}.

Here, we define two characters as interacting if they are directly
involved (\eg kiss, hug), or the actions of one influence the
actions of the other (\eg show something on a screen), or they
communicate (\eg talk to each other), or if they perform an activity
together (\eg shopping).  Two characters are not interacting within a
shot if they do not refer to each other (\eg both characters listen to
a third person talking), or they do not influence each other, or they
perform different tasks (\eg one character is watching TV while the
other is reading a book).

\subsection{Dataset processing and annotation}

\paragraph{Dataset. }
We use one episode of the TV show `Friends' (\textit{s03ep12}). 
First, we detect and track all heads (see Section~\ref{sub:d_t}), resulting in $1.7k$ head tracks. 
Then, with no further training, we apply \newnet on each track pair to determine if two characters are LAEO. 

\paragraph{Character annotation.}
All head tracks are annotated with the identity of their character. This results 
in main characters 
(more than one third of the tracks), irrelevant characters ($\sim35\%$), being wrong ($20\%$) or some secondary ones (the rest). 

\paragraph{Interaction annotation.}
Within each shot, all pairs of characters are annotated as interacting or not. 
Our annotation procedure results in $220$ positive and $200$ negative pairs.

\subsection{Experiments}
The goal is to assess whether LAEO can be used to predict character pair interactions at the shot level, and Friend-ness at the episode level. We measure LAEO at the shot level
using  `average-LAEO score' (AL) over the frames where
the two characters co-exist, and measure LAEO at the episode level as the average of AL over all shots in which the two
characters appear.
Interaction is a binary label for a pair of characters in a shot. We treat AL as the score for predicting interaction, and assess
its performance using Average Precision (AP).

\paragraph{Baselines.} 
For interaction prediction we use four baselines: 
(1)~Random Probability (PR): every pair has a random probability of interacting (drawn from an uniform distribution); 
(2)~Uniform Probability per Episode (UPE): the probability of interacting for a pair is $1/L$, where $L$ is the number of existing pairs per episode;  
(3)~Shots-Coexistence-Ratio (SCR): the ratio between the number of frames that two characters co-exist in a shot over the total number of frames of the shot; and
(4)~Uniform Probability per Shot (UPS): the probability of interacting for a pair is $1/L_{S}$, where $L_{S}$ is the number of existing pairs per shot.

\paragraph{Interaction prediction. }
The AP for individual pairs of characters is shown in Figure~\ref{fig:rankfriends2}(a) (more pairs in the suppl.\ material); 
and a pair-agnostic ranking, where all pairs are evaluated together, no matter the character identities is shown
in Figure~\ref{fig:rankfriends2}(b).

In Figure~\ref{fig:rankfriends2}(a), we observe that in some cases several baselines are good predictors as they capture the possible interactions, \eg \textit{ross-rachel} or \textit{monica-workmate monica}. However, in the cases where there exist several pairs within an episode or where two characters co-exist only in a few frames (compared to the shot length), the SCR (green) and UPS (blue) baselines are incapable of capturing the interactions, \eg \textit{joey-ross}, \textit{joey-monica} or \textit{ross-mark}. In these cases, however, AL (red bars) correctly predicts the interaction level between characters.  Overall, we observe that AL outperforms all other baselines in all pair-specific cases.

\begin{figure}[t!]
\begin{center}
\includegraphics[width=0.95\linewidth]{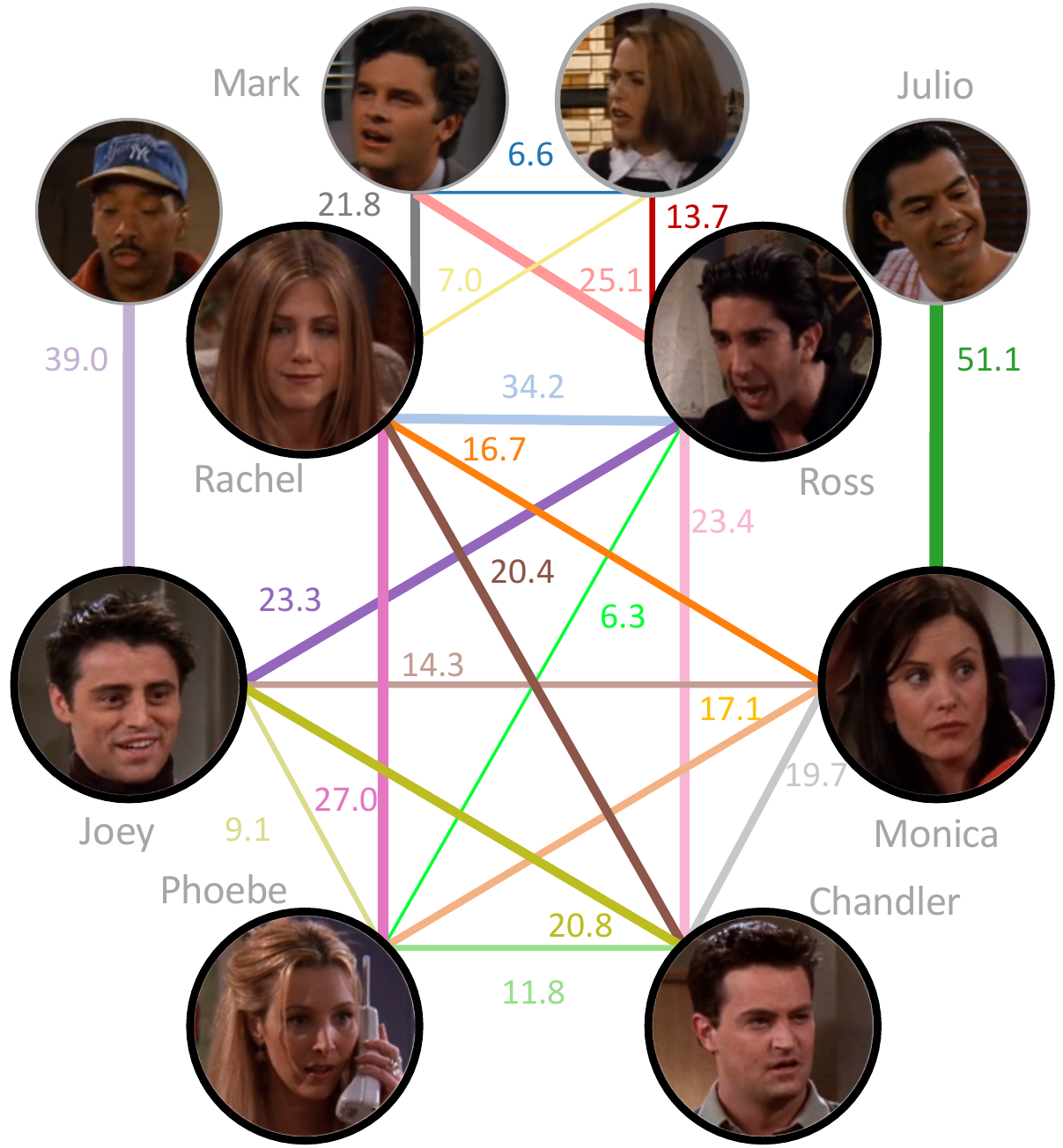}
\end{center}
 \caption{\small{\textbf{Social network using the \textit{Average-LAEO} (AL) on Friends.} 
 We depict the \%AL between character pairs with the edges in the graph: the thicker the edge, the more dominant the relationship. 
 We observe some clear patterns: Ross and Rachel or Monica and Julio `like' each other more than Chandler and Phoebe or Ross and Phoebe. 
 }}
 \label{fig:friends_examples}
\vspace{-4mm}
\end{figure}

In the pair-agnostic PR curves of Figure~\ref{fig:rankfriends2}(b), the AL score  outperforms all baselines by 4-34\%. 
The RP baseline performs worse than all other scores, which is expected as it contains no information, while UPE and SCR perform similarly (AP= 77\% and 68\%), indicating that the frequency of existence of a pair at the frame or episode level does not necessarily reveal interactions. 
The UPS score notably outperforms the other baselines by 3-29\% showing that the fewer people exist in a shot, the more likely they are to interact. 
Finally, AL outperforms all baselines, reaching AP=84\%; showing that it captures the main interactions with high confidence, and therefore can be useful for automatically retrieving them.

To demonstrate the powerfulness of AL and its superiority compared to SCR, we show some examples of pairs of characters in Figure~\ref{fig:rankfriends2}(c). 
The examples in green are correctly predicted as interacting by AL, but wrongly predicted as not-interacting by SCR; 
the examples in blue are correctly predicted as not-interacting by AL, but wrongly predicted as interacting by SCR;  
the examples in orange are missed interactions by AL, but correctly predicted as interacting by SCR. 
We observe that in several cases, the AL is suitable for predicting the presence or absence of interactions between characters, whereas the SCR is incapable of differentiating them; for instance, Monica and Joey in the last green example co-exist and interact in a few frames and, therefore, they are wrongly predicted as not-interacting by SCR. 
Moreover, we note that the AL fails to determine interactions where people are not LAEO (\eg Ross and Chandler or Mark and Rachel in orange). In most cases, however, either in real life or in TV-shows a human interaction typically involves gazing; hence, the AL is suitable for automatically capturing pairs of characters interacting.

\paragraph{\textit{Friend}-ness. }
For each shot, we measure \textit{friend}-ness between a pair of characters with the AL and depict it in the social network of Figure~\ref{fig:friends_examples}: the thicker the edge, the higher the score and the stronger the relations.  
AL captures the dominant relationships between characters, \eg Ross and Rachel, against characters that are more distant, \eg Phoebe and Chandler. 
Our study reveals all prominent pair relations, demonstrating that the more people are LAEO, the stronger their \textit{interaction} and \textit{social relationship}.

\section{Conclusions \major{and future work}} 
\label{sec:conclus}

In this paper, we focused on the problem of people \textit{looking at each other (LAEO)} in videos. 
We proposed \newnet, which takes as input head tracks and determines if the people in the track are LAEO. 
This is the first work that uses \textit{tracks} instead of bounding-boxes as input to reason about people on the whole track.  
\newnet consists of three branches, one for each character's tracked head and one for the relative position of the two heads.  
Moreover, we introduced two LAEO video datasets: \UCOLAEO and AVA-LAEO. 
Our experiments showed the ability of \newnet to correctly detect LAEO events and the temporal window where they happen. 
Our model achieves state-of-the-art results on the TVHID-LAEO dataset. 
Furthermore, we demonstrated the generality of our model by applying it to a social case scenario, where we automatically infer the \textit{social relationship }
between two people based on the frequency they LAEO \ie \textit{friend}-ness, and showed that our metric can be useful for guided search of interactions between characters in videos (\ie interaction prediction).
Finally, in Section 5 in the suppl.\  material we examine two other applications of \newnet, \ie head pose classification and regression. 
As future work, we identify the following research directions: incorporating explicit 3D information of humans (\eg \cite{zhu2017face}) to the model and exploring other kinds of social situations (\eg \cite{salsa}).
\ifCLASSOPTIONcompsoc
  \section*{Acknowledgments}
\else
  \section*{Acknowledgment}
\fi
We are grateful to our annotators (RF, RD, DK, DC, E. Pina), to Q. Pleplé for proof-reading, to S. Koepke for the model, to the reviewers for the constructive suggestions, and to NVIDIA for donating some of the GPUs we used. 
This work was supported by the Spanish grant ``Jos\'e Castillejo'', 
the EPSRC Programme Grant Seebibyte EP/M013774/1, and the Intelligence Advanced Research Projects Activity (IARPA) via Department of Interior/ Interior Business Center (DOI/IBC) contract \# D17PC00341.

\ifCLASSOPTIONcaptionsoff
  \newpage
\fi

{\small
\bibliographystyle{ieee}
\bibliography{shortstrings,laeo.bib}
}

\vspace{-5mm}
\vspace{-5mm}
\begin{IEEEbiography}[{\includegraphics[height=25mm,clip,keepaspectratio]{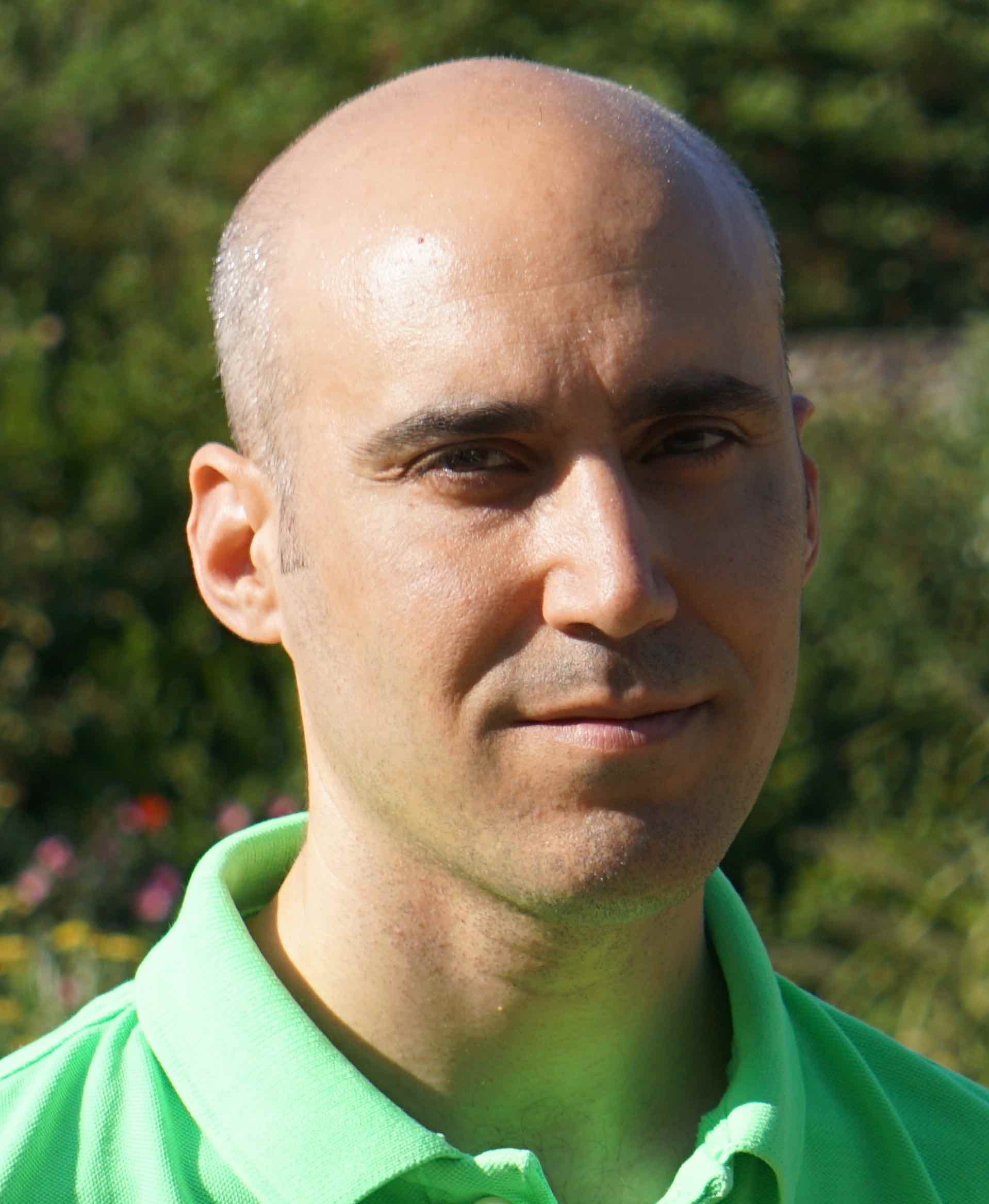}}]{Manuel J. Mar\'in-Jim\'enez}
received the BSc, MSc and PhD degrees from the University of Granada, Spain. 
He has published more than 60 technical papers at journals and international conferences. Currently, he works as Associate Professor at the University of C\'ordoba (Spain). 
His research interests include human-centric video understanding, gait recognition and continual learning. 
He is an active member of AERFAI, ELLIS and IEEE.
\end{IEEEbiography}

\vspace{-1.cm}
\begin{IEEEbiography}[{\includegraphics[height=25mm,clip,keepaspectratio]{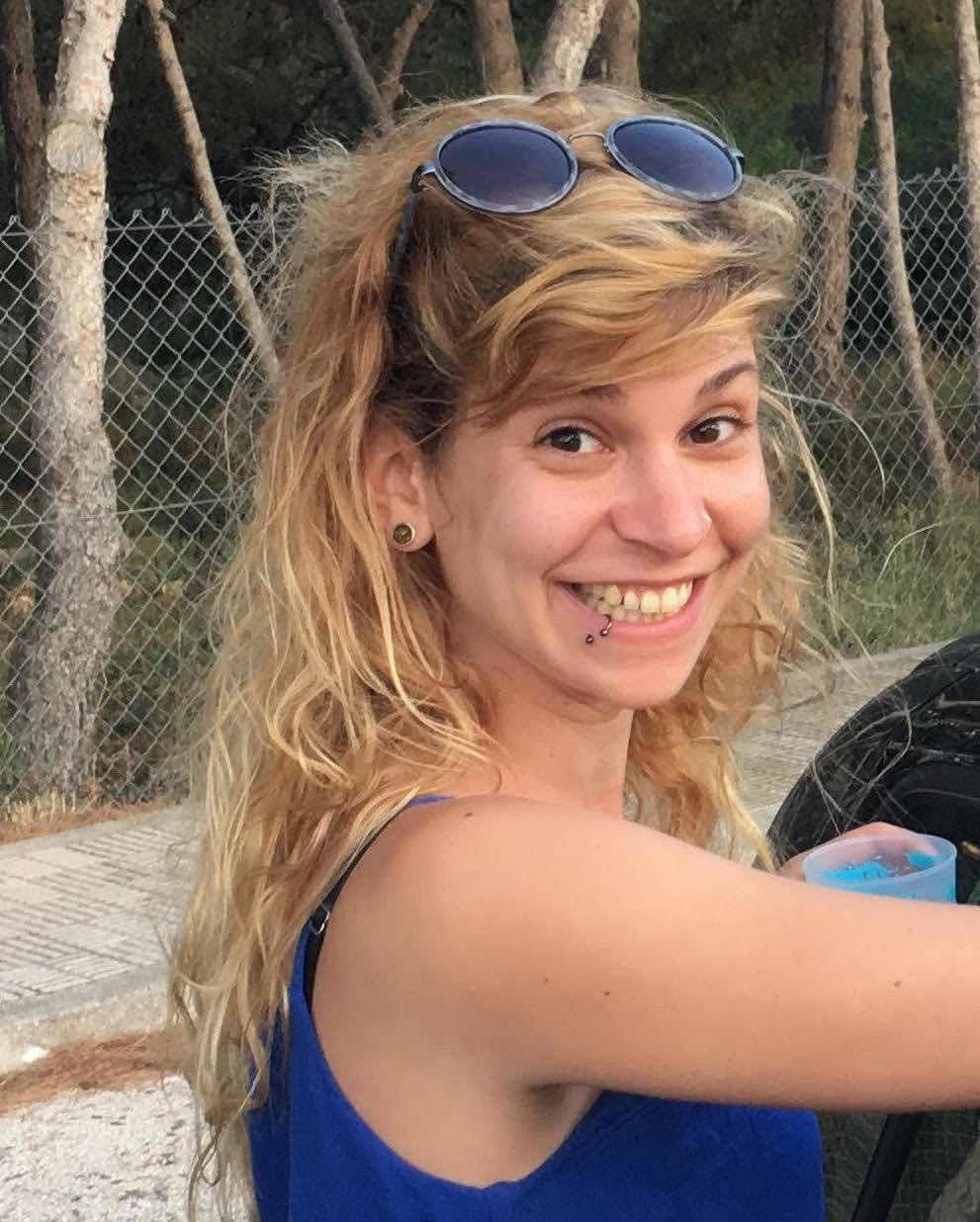}}]{Vicky Kalogeiton} is an Assistant Professor at LIX, École Polytechnique, IP Paris and a Research Fellow in the Visual Geometry Group of Oxford University since 2017. She received her PhD from the University of Edinburgh and INRIA Grenoble in 2017, and her M.Eng and M.Sc degrees from Duth, Greece in 2011 and 2013, respectively. Her research interests are computer vision and, specifically, human-centric video understanding.
\end{IEEEbiography}

\vspace{-1.cm}
\begin{IEEEbiographynophoto}{Pablo Medina-Su\'arez}
received the BSc degree from the University of C\'ordoba and is currently finishing his MSc studies. 
He collaborates with the Artificial Vision Applications group, at the University of C\'ordoba.
\end{IEEEbiographynophoto}

\vspace{-1.cm}
\begin{IEEEbiographynophoto}{Andrew Zisserman}
is a professor of computer vision engineering in the Department of Engineering Science, University of Oxford.
\end{IEEEbiographynophoto}



\section*{Supplementary material}

# Supplementary material
# LAEO-Net++: revisiting people Looking At Each Other in videos

Manuel J. Marín-Jiménez*, Vicky Kalogeiton*, Pablo Medina-Suárez, and Andrew Zisserman

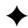

This supplementary material extends the main manuscript with (i) additional details about the architecture and training of LAEO-Net++ (Section 1); (ii) extended description of the head detection and tracking (Section 2); and (iii) more insights on the training loss function (Section 3). Then, we demonstrate (iv) some visually-ambiguous cases in the proposed AVA-LAEO dataset (Section 4), and (v) various applications of LAEO-Net++ in tasks different then LAEO detection (Sections 5-6).

## 1 LAEO-NET++ SPECIFICATIONS

### 1.1 LAEO-Net++ Architecture Specifications

As described in Section 3 in the main manuscript, LAEO-Net++ uses spatio-temporal 3D convolutions and can be applied to the head tracks in the video. It consists of three input branches, a fusion block, and a fully-connected layer. Two of the input streams determine the pose of the heads using five 3D conv laters, while the third represents their relative position and scale with four 3D conv layers. Table 1 reports the exact specifications of each conv and fully connected layer.

| head-pose<br>input: frame crops $64{\times}64{\times}\mathcal{T}$<br>(five 3D conv layers) | heads-map<br>input: map $64{\times}64$ $\times\mathcal{M}$<br>(four 3D conv layers) |
|---|---|
| f: $32 : 4 \times 4 \times 3$, s:$2 \times 2 \times 1$ | f: $16 \times 5 \times 5 \times 3$, s:$2 \times 2 \times 1$ |
| f: $64 : 4 \times 4 \times 3$, s:$2 \times 2 \times 1$ | f: $24 \times 3 \times 3 \times 3$, s:$2 \times 2 \times 1$ |
| f: $128 : 4 \times 4 \times 3$, s:$2 \times 2 \times 1$ | f: $32 \times 3 \times 3 \times 3$, s:$2 \times 2 \times 1$ |
| f: $256 : 4 \times 4 \times 3$, s:$2 \times 2 \times 1$ | f: $12 \times 6 \times 6 \times 1$, s:$1 \times 1 \times 1$ |
| f: $256 : 4 \times 4 \times 2$, s:$2 \times 2 \times 1$ | |

TABLE 1: **'Self-supervised style' architecture specification.** Branch details 'f': filter, 's': stride, $h \times w \times t$.

### 1.2 LAEO-Net++ training details

As mentioned in the paragraph 'Implementation details' of Section 8 in the main manuscript, LAEO-Net++ is implemented with

Keras [1] using TensorFlow as backend. Here, we provide more implementation details.

For implementing LAEO-Net++, we use zero-padding before the conv layers of the head-pose branches, batch normalization after the convolutions – with the exception of the first and last Conv layers, $\mu = 0.99, \varepsilon = 10^{-5}$ – and, leaky ReLu activations ($\alpha = 0.2$). For training it we use the Adam optimizer with mini-batches of 9 samples: 4 positives, 4 negatives and 1 hard negative. The learning rate starts at $10^{-4}$ and is reduced by a factor of 0.2 when the AP on the validation set does not improve for 10 consecutive iterations. The minimum possible learning rate is set to $10^{-7}$. Dropout is set to 0.2. Before applying the trained model to the test sets, the samples of the validation set of UCO-LAEO are added to the training set for few additional epochs using the latest learning rate.

## 2 HEAD DETECTION AND TRACKING

As mentioned in Section 5, LAEO-Net++ requires head tracks as input. Here, we describe in details the head detection and head-track generation.

**Head detection.** Our method requires head detections. In the literature, there are several models for face detection ([7, 8]); head detection, however, is a more challenging task, as it comprises detecting the whole head, including the face (if visible) but also the back of a head (e.g. [5]). We train a head detector using the Single Shot Multibox Detector (SSD) detector [4][1] from scratch. We train the model with a learning rate of $10^{-4}$ (first 50 epochs), and decrease it with a factor of 0.1 for the rest of the training. For speedup and better performance we use batch normalization and for robustness we use the data augmentation process from [4]. We train the head detector with the 'Hollywood heads' dataset [6]. It consists of head annotations for 1120 frames, split into 720 train and 200 val and test frames. We first train our detector with the training set and after validating the model, we train on the whole dataset as a refining stage.

**Head tracking.** Once we obtain head detections, we group them into tracks along time. For constructing tracks, we use the online linking algorithm of [2], as it is robust to missed detections and can generate tracks spanning different temporal extents of the video.

- (*) means equal contribution.
- Manuel J. Marín-Jiménez and Pablo Medina-Suárez are with the University of Córdoba, Spain.
  Emails: mjmarin@uco.es and i42mesup@uco.es
- Vicky Kalogeiton and Andrew Zisserman are with the University of Oxford.
  Emails: vicky@robots.ox.ac.uk and az@robots.ox.ac.uk

1. Detector: https://github.com/AVAuco/ssd_people

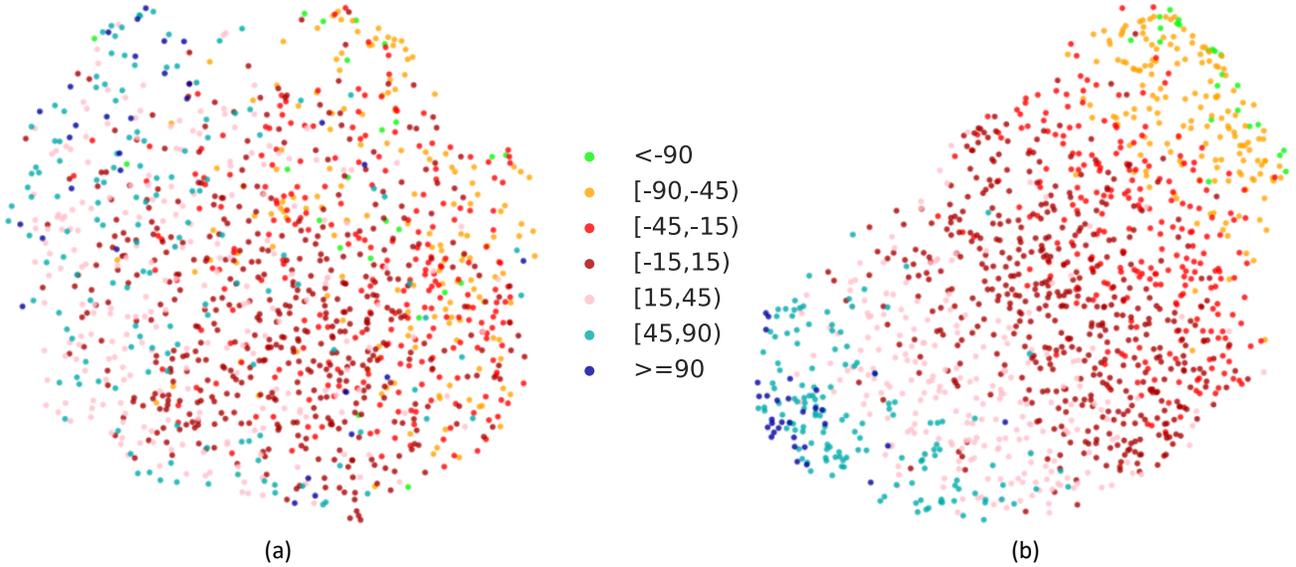

(a)                                                                                      (b)

Fig. 1: **Study of loss components.** UMAP projection of head embeddings using (a) just the $\mathcal{L}_s$ *'sign'* loss function (trained during 150 epochs) and (b) $\mathcal{L}_h$, i.e. combined 'sign+L1' (trained during just 15 epochs). We observe that $\mathcal{L}_s$ alone is not enough to distinguish head orientations, whereas the combined $\mathcal{L}_h$ clearly separates the head angles even though it is trained for only 15 epochs!

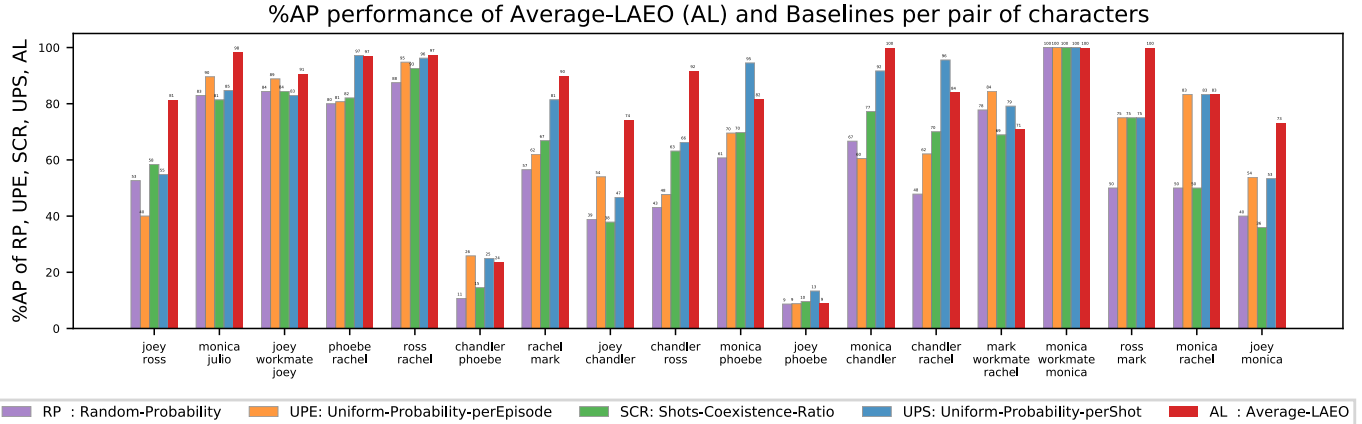

Fig. 2: **Interaction prediction with the *Average-LAEO* vs various Baselines on Friends.** In addition to Average-LAEO score (AL), we display four baselines: Random Probability (PR), Uniform Probability per Episode (UPE), Shots-Coexistence-Ratio (SCR), and Uniform Probability per Shot (UPS). (left) AP performance for the AL scores and various baselines for each pair, Some patterns are clear: Ross and Rachel or Monica and her workmate interact with each other almost continuously when they coexist; albeit their low frequency of co-existence, Joey and Ross, Joey and Monica or Ross and Mark interact significantly when they co-exist as captured mainly by AL (red). (*Best viewed in digital format.*)

Out of all head detections, we keep only the $N = 10$ highest scored ones for each frame. We extend a track $T$ from frame $f$ to frame $f + 1$ with the detection $h_{f+1}$ that has the maximum score if it is not picked by another track and ov $\left(h_f^T, h_{f+1}\right) \geqslant \lambda$, where ov is the overlap. If no such detection exists for $C$ consecutive frames, the track stops; otherwise, we interpolate the head detections. The score of a track is defined as the average score of its detections. At a given frame, new tracks start from not-picked head detections. To avoid shifting effects in tracks, we track both forwards and backwards in time.

## 3 LOSS FUNCTIONS TO TRAIN LAEO-NET++

Here, we examine the importance and impact of the proposed loss function (Eq. 3 in the main manuscript) for the fully-supervised pre-training scheme of LAEO-Net++. As a reminder, the loss function $\mathcal{L}_h$ for training the head-pose branch combines the individual losses for the yaw, pitch, and roll angles with a penalty term $\mathcal{L}_s$ for the yaw angle. In addition, all target angles are normalized (i.e. dividing them by the standard deviation of the orientations of the training set) before training.

**Importance of $\mathcal{L}_s$.** $\mathcal{L}_s$ penalizes an incorrect estimation of the *sign* of the yaw angle, i.e. failing to decide if the person is looking left or right. We demonstrate the importance of the sign of $\mathcal{L}_s$ by considering the following example. Let $\alpha$ represent the yaw angle. (i) For the pair $\alpha = 20$ and $\hat{\alpha} = 25$, the $L1$ value is 5. (ii) The same value ($L1 = 5$) is obtained for the pair $\alpha = -2.5$ and $\hat{\alpha} = 2.5$. In the former case, both the estimated and ground-truth angles indicate the same head orientation, but different magnitude.

In the latter case, however, one angle would indicate looking to the left, whereas the other would indicate looking to the right. This shows that distinguishing these cases is important for solving the LAEO problem and, therefore, using the additional 'sign' term of $\mathcal{L}_h$ is necessary.

**Impact of $\mathcal{L}_s$.** To demonstrate the impact of $\mathcal{L}_s$, we train and test LAEO-Net++ using only $\mathcal{L}_s$ (Eq. 2 in the main manuscript), with the same architecture of the head branch and the AFLW dataset. We obtain the corresponding head embeddings and display in Figure 1 (a) their UMAP projection. We observe that using only $\mathcal{L}_s$ results in mixed results: most angles are not clearly separated, but overall there exists a slight separation between extreme values of angles, e.g. angles more than 45 degrees (light and dark blue dots) are grouped towards the left part of the projection space, whereas angles between -90 and -15 degrees (orange and red dots) are mostly shifted towards the right part of the space. For comparison, we display in Figure 1 (b) the UMAP projection of the head embeddings when trained and tested with our full loss function $\mathcal{L}_h$ of Eq. 3 in the main manuscript. We observe that $\mathcal{L}_h$ results in a clear separation of the head angles, i.e. significantly better than the one from 'sign' alone even though it is trained for only 15 epochs.

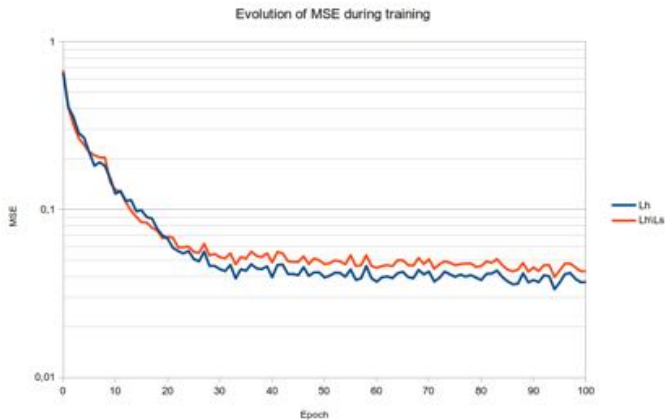

Fig. 3: **Contribution of $L_s$ to $L_h$.** Evolution of the Mean Squared Error (MSE) while training a head pose estimator: full $L_h$ loss vs $L_h$ without the $L_s$ loss. The lower, the better.

Finally, in addition to the previous experiments, we compare the contribution of the penalty term $L_s$ to the complete loss $L_h$. The plot in Figure 3 shows that the Mean Squared Error (MSE) obtained by the full loss function $L_h$ is lower than the one achieved by the ablated function $L_h \setminus L_s$. That means that the penalty term $L_s$ is helpful for the task.

## 4 AVA-LAEO DATASET

**Matching AVA annotations with head detections.** The original AVA dataset (v2.2) provides bounding boxes at the person-level. This means that a person bounding box may cover a full body, just an upper-body or even part of a head. Recall that our LAEO-Net++ uses as input head tracks. Therefore, in order to obtain head-level LAEO annotations, we match our head detections (and tracks) with the original bounding boxes of AVA. In particular, we define a (positive) match between a head detection and an

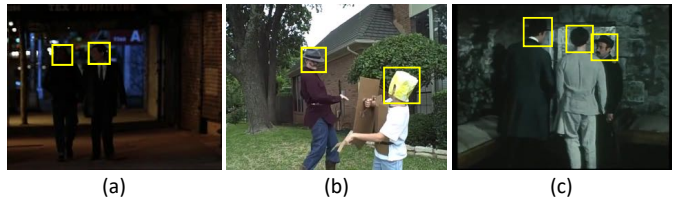

Fig. 4: **Samples annotated as 'ambiguous' by the annotators of AVA-LAEO.** For example, in (a) the background is too dark, in (b) the faces are not visible, and in (c) it is not *visually* clear which is the LAEO pair of the man on the right, if any, in the scene – our *human intuition* is that the most likely one is the most left man, but we cannot see his eyes, nor the eyes of the other man viewed from the back.

AVA bounding box if the intersection area over the head area is greater than a threshold. This requires that the head detection box is approximately within the AVA bounding box (as the latter is usually larger than the head one). This process results in a subset of people with their corresponding LAEO annotations.

**Ambiguous cases.** Figure 4 shows some typical visually ambiguous cases. These are due to blur or crowded scenes (c), no visible faces (b), dark background (a), and ambiguity due to the many possible combinations of mutual gaze (c).

## 5 FRIENDS EXAMPLES

In Figure 2 we include the %AP performance results for all pairs for the Average-LAEO (AL) and all other baselines. We observe that AL outperforms all other baselines. More details can be found in Section 9 in the main manuscript.

## 6 APPLICATIONS

In this section, we examine applications of LAEO-Net++ to other tasks. Specifically, we show that the weights learnt at the head branch of LAEO-Net++ for the LAEO detection problem can be transferred to two other tasks: (1) head orientation classification, and (2) head orientation regression.

For both experiments, we use the LAEO-Net++ obtained when training on AVA-LAEO with random inialization of the head branch, i.e. no pre-training. As a reminder, note that the size of each input head sample is $10 \times 64 \times 64 \times 3$ pixels (i.e. both training and test samples have had to be *inflated*). For both tasks, we use the AFLW dataset [3]. Specifically, we randomly select 1000 images for testing purposes, while making sure that the different yaw angles are approximately evenly distributed. Below, we describe them in details.

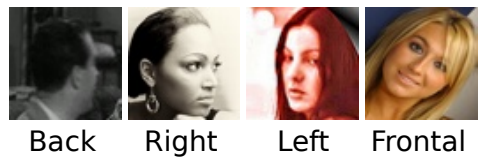

Back        Right        Left        Frontal

Fig. 5: **Classes defined for the 4-ways head pose classifier.** From left to right: near-back view, right, left and almost-frontal.

**Head orientation classification.** The goal of this task is to classify each head pose into four possible categories based on a discretization of the yaw angle. As shown in Figure 5, we define

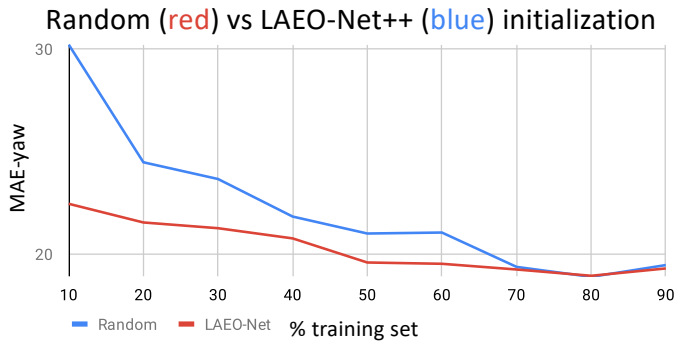

Fig. 6: **Training an estimator of the head yaw angle: different amount of training samples.** We compare the MAE (the lower the better) obtained on the test set for networks trained from scratch (i.e. random initialization of weights) against networks initialized from the head branch of a pretrained LAEO-Net.

four classes: back (absolute value greater than $120^o$), right, left and frontal (absolute value lower than $30^o$). The back view class is mainly represented by 210 samples extracted from the 'Hollywood heads' dataset [6]. Overall, the dataset for this task contains 23,495 samples (AFLW+back).

The proposed architecture for this task corresponds to the head branch of LAEO-Net++ plus an additional batch-normalization layer and a softmax layer with four units. The categorical cross-entropy loss function is used for training.

We investigate here a few-shot learning setup. We train during 5 epochs the proposed classifier with a subset of the training samples, i.e. 10% (2,290) and 50% (11,458), and we compare them with the results obtained by the same architecture initialized with random weights. The test accuracy is 60.9% (random) vs 68.2% (ours) for the 10% case, and 77.9% (random) vs 81.9% (ours) for the 50% case. These results suggest that the learnt weights are useful for this task.

**Head orientation regression.** This problem is more challenging than the previous one, as a real number has to be regressed. As in the previous task, we focus here on the yaw angle, i.e. turning head left-right. In contrast to the previous task, here we only include the annotations from AFLW with no additional back-view samples.

The architecture used for this task is composed of the LAEO-Net++ head branch plus a batch-normalization layer and a dense layer with a single unit (without any bias and activation function). We train the network with the L2 loss function.

Figure 6 summarizes the mean absolute error (MAE) in degrees obtained for different proportions of the training data, comparing the network obtained with random initialization against the network obtained from a pre-trained LAEO-Net++. As a reference, the 10% of the training data corresponds to 2,226 samples. In line with the findings of the previous experiment, initializing the regressor with LAEO-Net++ weights helps to speed-up the training process when only few samples are available.



\end{document}